\documentclass[journal,twocolumn]{IEEEtran}

\ifCLASSINFOpdf
  \usepackage[pdftex]{graphicx}
  \graphicspath{{./graphics/}}
  \DeclareGraphicsExtensions{.pdf,.jpeg,.png}
\else
  \usepackage[dvips]{graphicx}
  \graphicspath{{../graphics/}}
  \DeclareGraphicsExtensions{.eps}
\fi

\usepackage{cite}
\usepackage{amsmath}
\usepackage{array}
\usepackage{fixltx2e}
\usepackage{url}
\usepackage{hyperref}
\usepackage{enumitem}
\usepackage{subcaption}
\usepackage{pdflscape} 
\usepackage{multirow}
\usepackage{multicol}
\usepackage[flushleft,referable]{threeparttablex}
\usepackage{color}
\usepackage{flushend}
\usepackage{tablefootnote}
\usepackage{enumitem} 
\usepackage{todonotes} 
\usepackage{dblfloatfix} 
\usepackage{tabularx} 
\usepackage{colortbl}
\usepackage{booktabs}
\usepackage[bottom]{footmisc}
\usepackage{footnote}

\newcommand{\ie}{{\it i.e.},~}
\newcommand{\eg}{{\it e.g.},~}
\newcommand{\etal}{{\it et al.}~}

\usepackage[normalem]{ulem}

\colorlet{tableheadcolor}{blue!25} 
\colorlet{tablerowcolor}{blue!10} 
\newcommand{\topline}{\arrayrulecolor{black}\specialrule{0.1em}{\abovetopsep}{0pt}%
    \arrayrulecolor{white}\specialrule{\belowrulesep}{0pt}{0pt}%
    \arrayrulecolor{black}}

\begin{document}

\title{Ensemble of Multi-View Learning Classifiers for Cross-Domain Iris Presentation Attack Detection\\
}

\author{Andrey Kuehlkamp,~\IEEEmembership{Student Member,~IEEE,}
        Allan Pinto,~\IEEEmembership{Student Member,~IEEE,}\\
        Anderson Rocha,~\IEEEmembership{Senior Member,~IEEE},
        Kevin W. Bowyer,~\IEEEmembership{Fellow,~IEEE,}
        Adam Czajka,~\IEEEmembership{Senior Member,~IEEE}
\thanks{A. Kuehlkamp, K. Bowyer and A. Czajka  are with the University of Notre Dame, USA. E-mail: \{akuehlka,kwb,aczajka\}@nd.edu}
\thanks{A. Pinto and A. Rocha are with the Institute of Computing, University of Campinas (Unicamp), Av. Albert Einstein, 1251, Campinas, SP, Brazil, 13083-852. E-mail: \{allan.pinto,anderson.rocha\}@ic.unicamp.br.}
\thanks{Manuscript received ...; revised ...}}

\maketitle

\markboth{Draft}{Draft}

\begin{abstract}
The adoption of large-scale iris recognition systems around the world has brought to light the importance of detecting presentation attack images (textured contact lenses and printouts).
This work presents a new approach in iris Presentation Attack Detection (PAD), by exploring combinations of Convolutional Neural Networks (CNNs) and transformed input spaces through 
binarized statistical image features (BSIF). Our method combines lightweight CNNs to classify multiple BSIF views of the input image. Following explorations on complementary input spaces leading to more discriminative features to detect presentation attacks, we also propose an algorithm to select the best (and most discriminative) predictors for the task at hand. An ensemble of predictors makes use of their expected individual performances to aggregate their results into a final prediction. Results show that this technique improves on the current state of the art in iris PAD, outperforming the winner of LivDet-Iris 2017 competition both for intra- and cross-dataset scenarios, and illustrating the very difficult nature of the cross-dataset scenario.
\end{abstract}


\ifCLASSOPTIONpeerreview
\begin{center} \bfseries EDICS Category: BIO-MODA-IRI\end{center}
\fi

\IEEEpeerreviewmaketitle


\section{Introduction}

\IEEEPARstart{H}{ow} can we distinguish between two individuals without reasonable doubt? This question has motivated biometric researchers for centuries --- from the pioneer works of Bertillon and Galton
to recent advances in biometric-enabled mobile payments and wearable devices.
Numerous biometric characteristics emerged, and sometimes faded away, as the field progressed, but one biometric trait that has surely withstood the test of time is iris recognition. The iris pattern is unique and ``determined epigenetically by random events in the morphogenesis of this tissue'' \cite{Daugman_PRS_2001}, and thus offers high discrimination power, making it useful in distinguishing even identical twins~\cite{Bowyer_BTAS_2016}.

The recognition power and matching speed of iris recognition has propelled it into use in large-scale applications, \eg Unique ID in India~\cite{UIDAI,Daugman:SPIE:2014}, and the NEXUS system operated jointly by the Canada Border Services Agency and the U.S. Customs and Border Protection to speed up the identification of pre-screened travelers \cite{NEXUS_URL}. As iris recognition becomes more pervasive, the number and variety of attempted attacks naturally intensify, and the problem of \emph{presentation attack detection} (PAD) becomes an essential research topic.

According to the standardized vocabulary in ISO/IEC 30107-1, a \emph{presentation attack} is a ``presentation to the biometric data capture subsystem with the goal of interfering with the operation of the biometric system'' \cite{ISO_30107-1_2016}. An impostor may seek to obtain unauthorized access to an authentication system by \emph{impersonating} a legitimate user, or may want to intentionally \emph{conceal} his or her identity to avoid recognition. Presentation attacks can be realized in various ways, in particular by presenting artifacts, such as paper printouts or textured contact lenses, non-conformant use of a biometric sensor, or even presenting cadaver eyes to the sensor. In this paper, we focus on attacks related to presentation of artificial objects.
\begin{figure}[!tb]
    \centering
    \begin{subfigure}[b]{0.99\linewidth}
        \includegraphics[width=\linewidth]{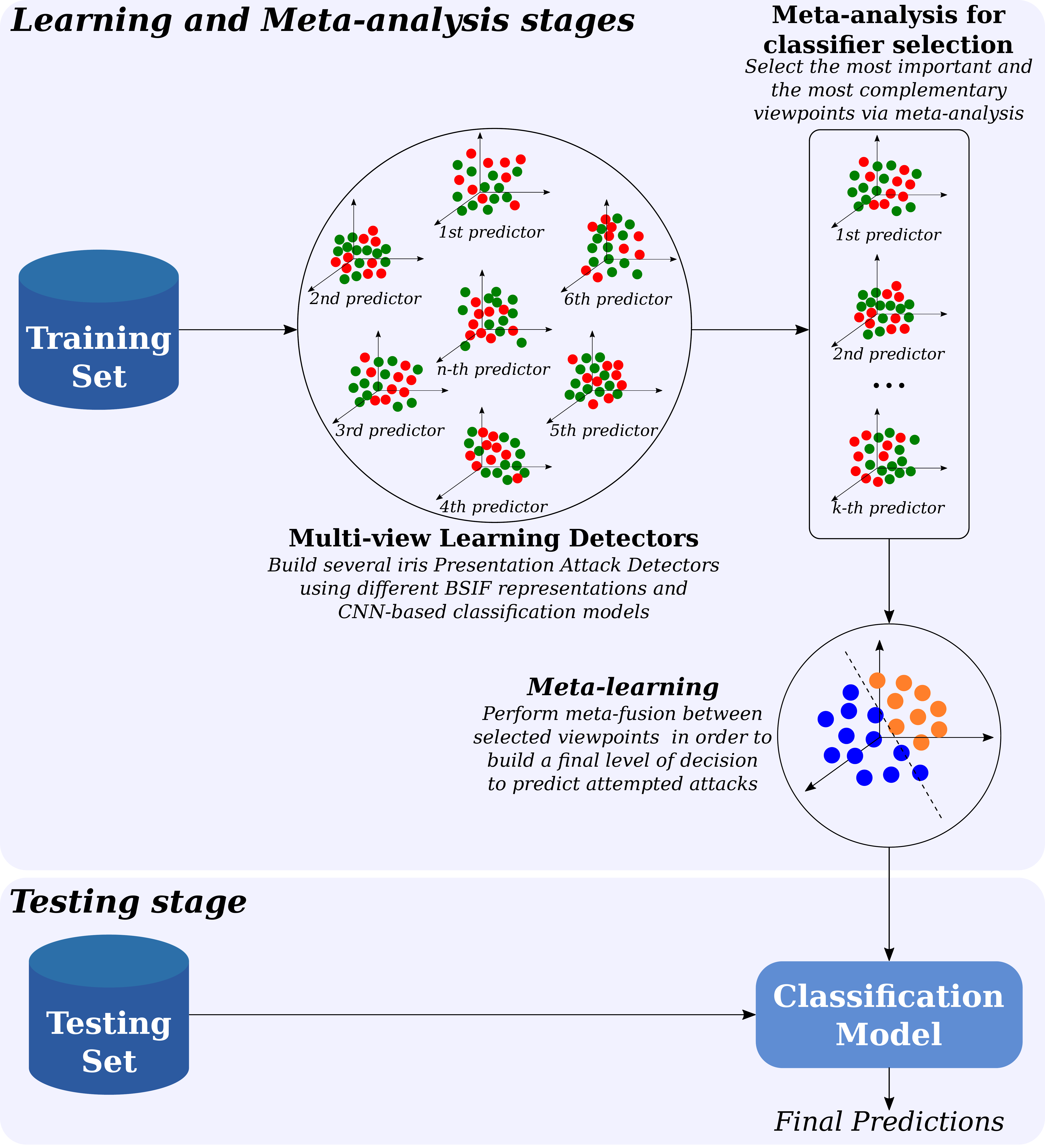}
    \end{subfigure}
    \caption{Overview of the proposed method. The first step is generating multiple views from dataset by training lightweight CNNs fed with different BSIF representations, referred to as multi-view-CNN predictors. Next, we select the most promising predictors according to their relevance and complementarity, and combine them via meta-fusion approach. 
    }
    \label{fig:method_overview}
\end{figure}

The Iris Liveness Detection Competition (LivDet-Iris, \url{www.livdet.org})
was begun in 2013.
The third and the most recent edition took place in 2017~\cite{Yambay2017}.
This competition is focused on properly measuring how well current technology withstands presentation attacks employing artifacts.
The sequence of competitions provides an important insight into 
the pace of evolution of iris PAD methods. 
The results reported in the 2017 competition show that iris PAD algorithms are still far from achieving acceptable detection rates. Moreover, the LivDet 2017 results suggest that challenging evaluation protocols, such as cross-dataset and cross-sensor setups --- hereinafter referred to as cross-domain --- can be considered the major limitation of current PAD algorithms and a current open research problem.

In the face of the evident need for better iris PAD,
this paper introduces a new technique based on exploiting multiple transformations of the input data so as to enhance complementary patterns, leading to a more discriminant manifold separating genuine authentication attempts from attacks.
The multiple transformations are obtained through the combination of hand-crafted and data-driven approaches. While Binarized Statistical Image Features (BSIF)~\cite{Kannala_ICPR_2012} is a popular, effective and partially hand-crafted texture descriptor, Convolutional Neural Networks (CNN) complement the repertoire with powerful description methods capable of learning even the subtlest discrimination clue present in the available training data through a series of non-linear transformations on the input data. 

Although BSIF- and CNN-based methods have been used before in iris PAD, to our knowledge there are no published papers dealing with the challenges of cross-domain deployments, or exploiting complementary properties of various feature extractors and classification strategies when defining a discriminative manifold for the problem under cross-domain constraints. In this vein, we introduce a way of combining both methodologies, hand-crafted and data-driven, in order to offer an iris PAD algorithm that better generalizes to unknown attack types. In addition, we also present a novel meta-analysis algorithm for selecting and aggregating the most prominent data views (\ie transformations), based on two well-known techniques for feature selection: the random forest importance feature weighting and the inter-rater agreement measures, so as to provide the most accurate detection method with the least possible computational impact. Fig~\ref{fig:method_overview} gives an overview of the proposed method.

In summary, the main contributions and novelties of this work are:
\begin{itemize}
    \item{A new approach that leverages multiple pre-trained BSIF filters to effectively train lightweight CNNs;}
    \item{A new fusion algorithm that selects and combines multiple classifiers, considering their importance and complementarity;}
    \item{A thorough cross-domain evaluation of the problem on datasets currently used to document the state of the art in the field;}
    \item{A new PAD method that outperforms the winner of the most recent LivDet-Iris competition, the authoritative international challenge on the subject.}
\end{itemize}

To encourage reproducibility, the source code of our implementation will be publicly available on GitHub. The datasets are already available through the LivDet competition. In the remainder of this paper, we briefly survey important iris PAD methods in prior art (Sec.~\ref{sec:related_work}) and introduce the proposed methodology (Sec.~\ref{sec:proposed_method}). Then we present experiments and validation (Sec.~\ref{sec:experimental_results}) and, finally, draw conclusions and present possible future work (Sec.~\ref{sec:conclusions}).

\section{Related Work}
\label{sec:related_work}

The first ideas for countermeasures against iris presentation attacks were proposed some 18 years ago by Daugman \cite{Daugman_IMAIP_2000} and became a basis of many current effective PAD methods. Early Daugman's concepts include finding anomalies in Fourier spectrum to detect printed irises, either on a paper or on a contact lens, detection of specular ``Purkinje'' reflections from both the cornea and the lens, or investigating pupil size variations, either spontaneous (``hippus'') or stimulated by visible light.

In general, there are two goals in presentation attacks, impersonation or identity concealment. The first successful demonstration of impersonation with the use of a commercial sensor was shown by Thalheim \etal \cite{Thalheim_CT_2002}. They used iris images printed on a paper, with a hole cut where the pupil was printed, to make a successful impersonation attack on a commercial iris recognition system with authentic eyes previously enrolled. The first use of one's eye to evade recognition observed in an operational environment was recorded at the border crossing point employing iris recognition in the United Arab Emirates \cite{Al-Raisi_TI_2008}. The attackers administered eye drops to make the pupil excessively dilated. This made the iris texture deformed too severely to be compensated by feature extraction and matching algorithms, and hence generating false non-matches. Since these early demonstrations of vulnerabilities, other presentation attack instruments have been studied, including use of textured contact lenses that partially occlude the actual iris texture \cite{Doyle_ICB_2013}, presentation of iris images displayed on a screen \cite{HeXiaofu_ICB_2009}, or use of prosthetic eyes \cite{Zuo_TIFS_2007}. After recent studies by Trokielewicz \etal \cite{Trokielewicz_BTAS_2016} presenting that post-mortem iris recognition is possible for a couple of weeks after death, cadaver eyes can also be considered as a potential presentation attack instrument.

%
\begin{figure*}[!tb]
    \begin{subfigure}[b]{0.32\linewidth}
            \includegraphics[width=\linewidth]{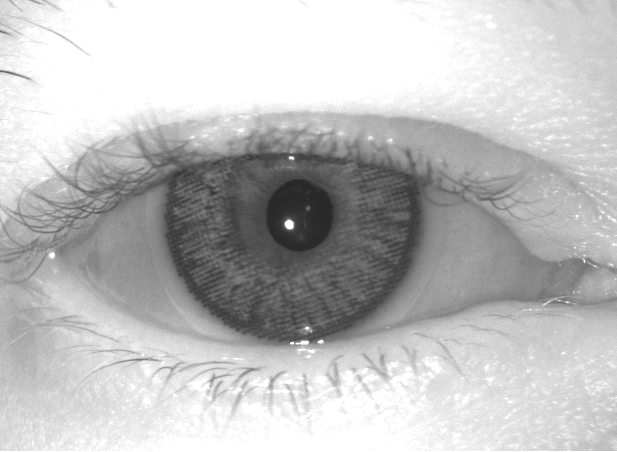}
        \caption{Raw iris image}
        \label{fig:origimg}
    \end{subfigure}
    ~
    \begin{subfigure}[b]{0.32\linewidth}
        \includegraphics[width=\linewidth]{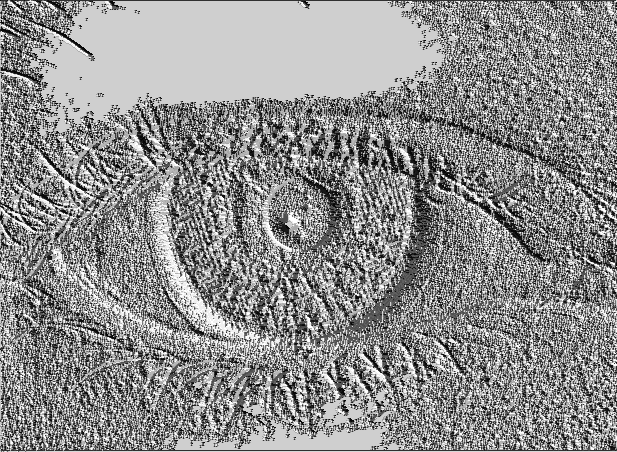}
        \caption{$B_{3\times 3 \times 5}$}
        \label{fig:bsif3x3}
    \end{subfigure}
    ~
    \begin{subfigure}[b]{0.32\linewidth}
        \centering
        \includegraphics[width=\linewidth]{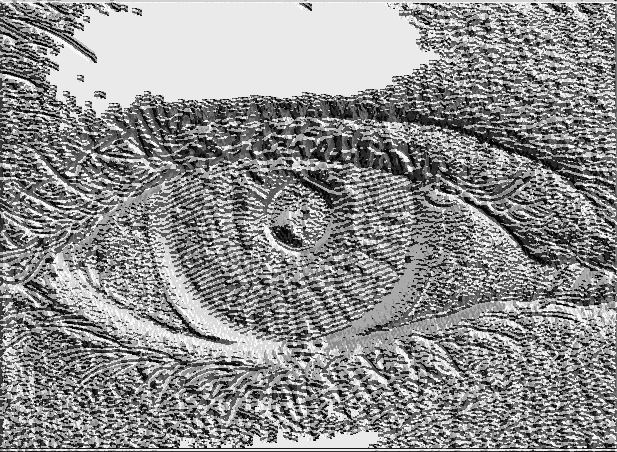}
        \caption{$B_{7\times 7 \times 12}$}
        \label{fig:bsif7x7}
    \end{subfigure}
    \caption{Examples of different BSIF representations $B_{n\times n \times l}$ (\ref{fig:bsif3x3},\ref{fig:bsif7x7}) of the same image (\ref{fig:origimg}). Note how the structure of the contact lens is highlighted through the different representations.}
    \label{fig:bsif_examples}
\end{figure*}

There is a rich literature on PAD methods presenting various levels of sophistication, and putting different requirements on the sensors' configuration and signals necessary to detect presentation attacks. In the largest group of PAD methods, a single near-infrared iris image, compliant to ISO/IEC 19794-6, is used in both identity verification and presentation attack detection. The \emph{hand-crafted} approaches use various image descriptors to calculate image features, which are used to distinguish between authentic irises and artifacts typically through the use of Support Vector Machine classifiers. Popular techniques used in calculation of PAD-related iris image features are Binarized Statistical Image Features (BSIF) \cite{Komulainen_IJCB_2014}, Local Binary Patterns (LBP) \cite{Doyle_ICB_2013}, Binary Gabor Patterns (BGP) \cite{Lovish_CAIP_2015}, Local Contrast-Phase Descriptor (LCPD) \cite{Gragnaniello_TIFS_2015}, Local Phase Quantization (LPQ) \cite{Sequeira_TSP_2016}, Scale Invariant Descriptor (SID) \cite{Gragnaniello_SITIS_2014}, Scale Invariant Feature Transform (SIFT) and DAISY \cite{Pala_CVPR_2017}, Locally Uniform Comparison Image Descriptor (LUCID) and CENsus TRansform hISTogram (CENTRIST) \cite{Akhtar_AVSS_2014}, Weber Local Descriptor (WLD) \cite{Gragnaniello_TIFS_2015}, Wavelet Packet Transform (WPT) \cite{Chen_PRL_2012} or image quality descriptors proposed by Galbally \etal \cite{Galbally_Handbook_2016}. Instead of ``hand-crafting'' effective feature extractors, one may also benefit from recently popular \emph{data-driven} approaches that learn directly from the data how to process and classify iris images to solve the PAD task  \cite{Silva_SIBGRAPI_2015,Menotti:TIFS:2015,Gragnaniello_SITIS_2016,He_BTAS_2016,Pala_CVPR_2017,Raghavendra_WACV_2017}. 

The above PAD methods rely upon an iris image that is typically used later for biometric recognition, and hence they can be implemented in the existing iris recognition sensors. However, if some hardware adaptations are possible, and some more complicated static features of the eye can be measured, one may consider multi-spectral analysis \cite{Lee_BS_2006,Park_OptEng_2007,Chen_PRL_2012,Thavalengal_TCE_2016} or estimation of three-dimensional iris features \cite{Pacut_ICCST_2006,Lee_IMA_2010,Connell_ASSP_2013,Hughes_HICSS_2013} as potential PAD techniques. Making the PAD more complex, one may consider measuring spontaneous dynamic features of the eye, such as micro-movements of an eyeball, either using Eulerian video magnification \cite{Raja_TIFS_2015} or by using an eye-tracking device \cite{Rigas_PRL_2015}. If there is a possibility to additionally stimulate the eye with varying visible light, and measure its reaction, the use of pupil dynamic models may help to easily detect static or oddly-behaving artifacts \cite{Czajka_TIFS_2015,Thavalengal_TCE_2016}. A recent survey  by Czajka and Bowyer provides a comprehensive assessment of the state of the art in iris PAD \cite{Czajka_CSUR_2018}.

In \cite{Nguyen_2018}, a PAD approach based on deep features extracted from different iris regions and classified by feature and score-level fusion of SVM is described. The authors report very low error rates on a subset of the LivDet-Iris 2017 datasets.
Kontschieder \etal \cite{kontschieder2015deep} published a work that seeks to combine deep convolutional networks for feature extraction, with the classification power of Random Forests. They describe an alternative approach to train Random Forest classifiers, which consists of a stochastic version of decision trees that are trainable through backpropagation. Random Forests trained with this method can either be standalone classifiers or act as alternative classifiers on top of a CNN. The authors claim to have outperformed the state of the art in image classification when integrating it to a GoogLeNet network.

Evaluation of PAD reliability significantly differs from statistical evaluation of biometric recognition. ISO/IEC JTC1 subcommittee 37 issued both the PAD-related vocabulary \cite{ISO_30107-1_2016}, and recommendations on how to evaluate and report the PAD-related performance \cite{ISO_30107-3_2017}. An important effort related to iris PAD evaluation is the LivDet-Iris competition series (\url{http://livdet.org/}), which has had editions in 2013 \cite{Yambay2014}, 2015 \cite{Yambay2015} and 2017 \cite{Yambay2017}. The  2017 edition of LivDet-Iris is the most recent, global, independent evaluation of PAD algorithms for detection of iris printouts and textured contact lenses. This paper follows exactly the LivDet-Iris 2017 evaluation protocol, and the results presented herein are directly comparable with the LivDet-Iris 2017 winning solution.
\section{Proposed Method}
\label{sec:proposed_method}

The task of the PAD algorithm is to capture differences between an actual live iris and either non-iris object, non-conformant use of an actual iris, or a cadaver eye. In the simplest scenario, when only static features of a single iris image are used, the PAD method recognizes anomalies in the presented iris texture. However, given different fabrication processes and the richness of details in an iris, it is likely that any single texture descriptor cannot capture all necessary leads hinting at a possible attack. Therefore fusions of different texture descriptors (\eg LBP, BSIF, Gabor filters) and classifiers (\eg Support Vector Machines, Neural Networks, etc.) have also been explored in prior literature.

Although texture analysis has been a staple in iris research over the years, development of image processing methods that capture such intricate textures has been a challenge. Therefore, more recently some researchers have brought to bear data-driven techniques, especially deep CNNs, to learn directly from training data the 
iris features useful in PAD. Normally, the input of such networks are the raw pixels themselves. Even though such data-driven methods have led to good results, these models do not work well in the so-called cross-domain setup, \ie when different training/testing conditions are part of the problem. As such, the conditions during training are not enough to allow robust generalization during testing.

This is where our work herein comes into play. In this section, we present a way of accelerating and empowering CNNs to capture texture patterns in such a way that differences among genuine and attack samples can be more easily spotted. Conscious of the representational power of CNNs to properly learn discriminative features, but at the same time of their innate need of large amounts of training data, we facilitate the process of learning by feeding the network with transformed input that highlights texture features important to make a distinction between a live iris and a presentation attack iris. We achieve this by transforming the input image into Binarized Statistical Image Features (BSIFs)~\cite{kannala2012bsif}. Second, 
it is likely that looking at the iris texture patterns from different vantage points, including different scales, might allow us extracting more features capturing differences between authentic iris patterns and artifacts. 

Thus, we consider using multiple BSIF filter sets, characterized by a scale $l$ and a depth $n$ (number of filter kernels in a single set) to create BSIF representations $B_{n\times n \times l}$ for each $(i,j)$ pixel of the original image $I$:

$$
B_{n\times n \times l}(i,j) = \sum_{k=0}^{n-1}b_k(i,j)2^k
$$

where

$$
b_k(i,j) =  
\begin{cases}
    1 & \text{if } s_k(i,j) \geq 0\\
    0 & \text{otherwise}
  \end{cases},
$$

$$
s_k(i,j) = \sum_{u=0}^{l-1}\sum_{v=0}^{l-1} w_{k;n,l}(u,v) I(i+u,j+v),
$$

\noindent
and $w_{k;n,l}$ is the $k$-th filter kernel in the set of filters derived for a given $n$ and $l$. Using original sets of BSIF filters, as proposed by Kannala and Rahtu~\cite{kannala2012bsif}, allows calculating 60 different BSIF representations of a single image. Each of these 60 image transformations is then used by one CNN to learn higher-level features for discriminating authentic and presentation attack irises. Finally, with different learned features at hand, a natural question is how to effectively select the best ones for PAD detection while eliminating the non-representative ones. For this we exploit two different fusion schemes: one based on random-forest feature weighting and meta-learning strategies, and the second relying upon inter-rater agreement measures.


BSIF kernels included into each of 60 sets were calculated from natural images in a way that maximizes statistical independence of filter responses, and the Independent Component Analysis was used for this purpose \cite{kannala2012bsif}. Assuming that patterns observed in artifacts are statistically independent of textures observed in authentic irises, such decomposition of images, and then building BSIF representations of image based on binarized filter responses, may facilitate calculating PAD-related features from BSIF representations instead of raw images.

While it is theoretically possible to train a CNN to obtain the same filters as BSIF in its first convolutional layer, the cost function and the training process would be significantly different from our goals and would likely require more training data than one normally has for solving a PAD problem. To reduce the amount of training data, we thus transform the images to a new space represented through BSIF operations and this new input serves as an additional transformation layer for the CNNs. As we go toward the output layers in the CNN, such features are further specialized to higher-level representations. The idea is that feeding the network with a transformed input would allow it to more quickly achieve such deeper representations without extensive training. However, some useful features may still not be learned from this transformed space. Therefore, drawing on good results from previous studies in the PAD literature~\cite{Menotti:TIFS:2015, livdet2017}, in addition to learning features from these transformed spaces, we also consider learning features directly from raw image inputs. At the end, with fusion strategies, we can show how such different treatments are complementary and how some of them do not contribute to separate authentic samples from artifacts. The key aspect of our methodology is that different filters learned from transformed spaces capture richer details than just a single representation and translate to better results in the difficult cross-dataset setup, the one in which training and testing conditions are different. 

Finally, we end up with 61 CNN-based predictors: 60 fed by BSIF representations and 1 fed by the raw image. We refer to such complementary CNNs learned with different input representations as multi-view-CNN predictors. Finding the optimal decomposition of the numerous possible configurations of the BSIF filter sets is not straightforward. That is exactly why the combination of CNNs is powerful. 

Since one of the CNN-based predictor operates on a raw image, it is interesting to compare its first-layer filters with those used in BSIF transformation. Visual comparison of these filters reveals some similarities: at a small scale ($3 \times 3$), most of them resemble edge, corner or dot detectors, similarly to what we would expect in the V1/V2 regions of the brain specialized in roughly describing visual inputs through edges and corners. However, BSIF $3 \times 3$ filters and CNN $3 \times 3$ filters are different. This means that the CNN operating on raw images developed its own way to preprocess the images in the first layer when compared to BSIF-based transformation, which may suggest that this predictor is complementary to the remaining 60 predictors, and it is worth using in the fusion.

Fig.~\ref{fig:method_overview} illustrates the main steps in our approach, which are explained in details in the following sections.

\subsection{Feature Extraction}

We compute BSIF representations by exploiting filter width size ranges from $3\times3$ to $17\times17$, and depth (\ie number of filters in a set) ranges from $5$ to $12$. Fig.~\ref{fig:bsif_examples} shows some examples of BSIF views of the same image presenting an iris with a textured contact lens. Note how different BSIF representations highlight the pattern of a contact-lens.

We use OSIRIS~\cite{othman2016osiris} (version 4.1)~\footnote{Source code is available in \url{http://svnext.it-sudparis.eu/svnview2-eph/ref_syst/Iris_Osiris_v4.1/}} to find the center of the iris in the original (not BSIF-transformed) image. Next, we crop a $260 \times 260$ region around the iris in both the BSIF-transformed images and the original image. These cropped samples are input to the CNNs in the next step. This $260 \times 260$ size is based on the size of iris images in commercial sensors ($640 \times 480$) and the ISO/IEC 19794-6 recommendation for a minimum 120 pixels across the iris diameter.


%
\begin{figure*}[!htb]
    \centering
    \includegraphics[width=\linewidth,
                     trim=0.5cm 0.5cm 0.6cm 0cm]{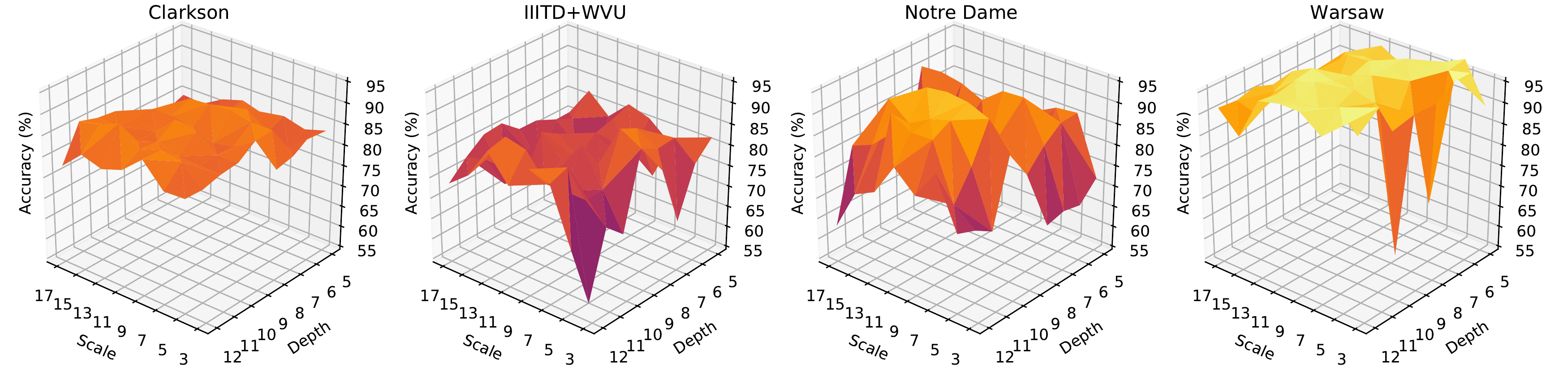}
    \caption{Performance comparison of the proposed CNN operating on BSIF inputs of varying scale and depth. Accuracies are estimated over the \emph{test unknown} partition of each dataset.}
    \label{fig:bsif_analysis}
\end{figure*}

To substantiate the descriptive advantage offered by BSIF filtering over the raw image input, we performed a baseline classification using a linear SVM to compare classification accuracies. 
While the best accuracy obtained with raw images was below 80\%, some of the BSIF filters achieved performance superior to 95\%. 
Besides, the mean accuracy of BSIF images was higher, confirming the potential benefit of  some of the filters.

With the intention of better understanding the results of BSIF images as input features, we performed an ablation analysis on the effect of scale and depth of the BSIF filters on the final classification. A CNN classifier (described in Section \ref{sec:our_cnn}) was trained on different input images, generated by BSIF filtering, first varying filter sizes, and then their depths. Results of this analysis are shown in Figure~\ref{fig:bsif_analysis}. It is possible to observe that some particular scales or depths have some (positive or negative) impact on the classification accuracy, but it is not possible to infer a generalizable trend.

\begin{figure}[!htb]
    \centering
    \includegraphics[width=0.98\linewidth,
                     trim=0.6cm 0.5cm 0.5cm 0.4cm]{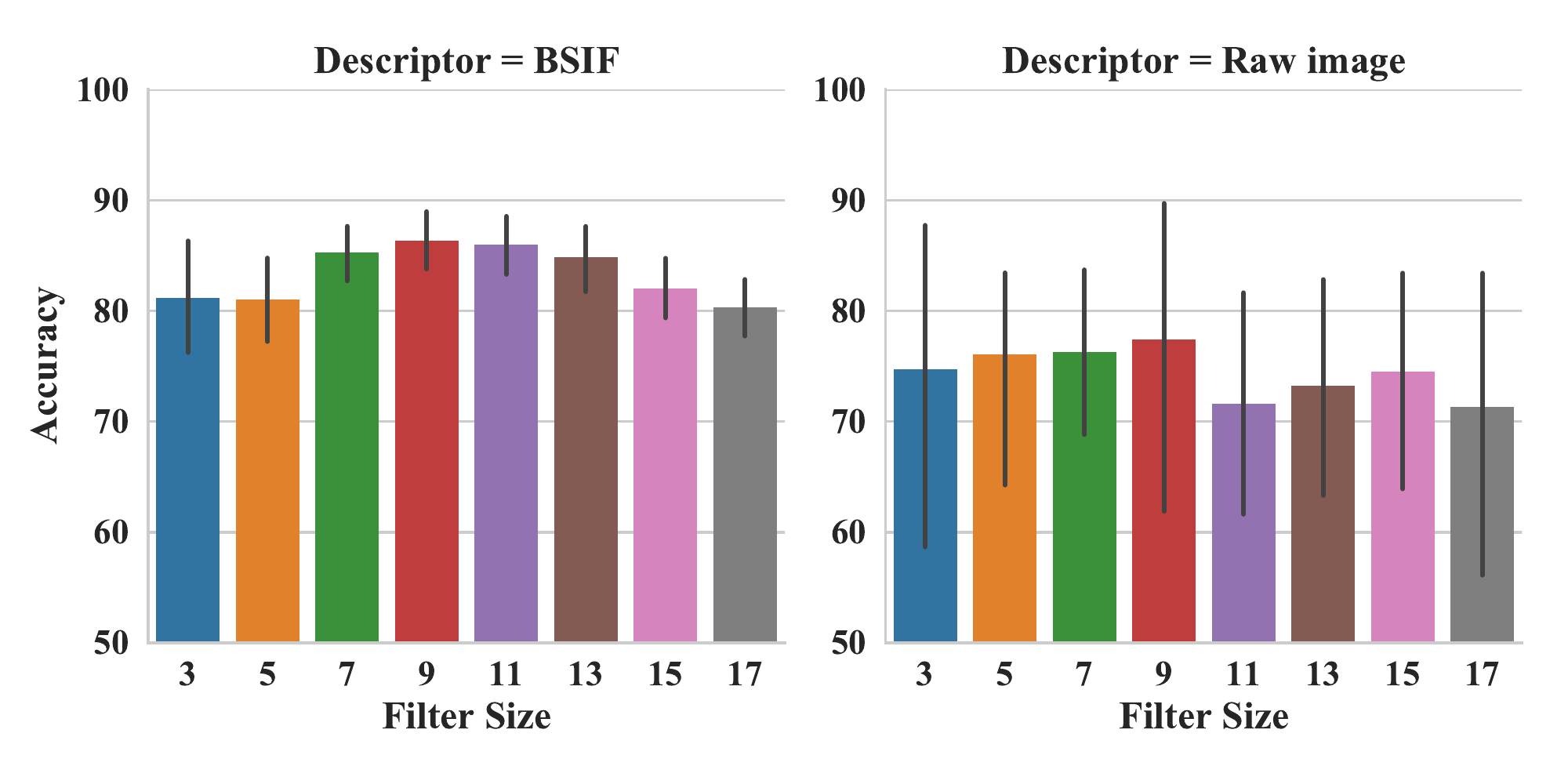}
    \caption{Performance comparison between BSIF and raw image as the input to the CNN. Accuracies are estimated over the \emph{test unknown} partition of each dataset. The length of the whiskers equals to one standard deviation of the obtained results.}
    \label{fig:cnn_analysis}
\end{figure}

In the same vein, we also tried to vary the size of the filter in the first convolutional layer of the CNN, while training it to perform classification on original images, and compared these results with BSIF (Fig. \ref{fig:cnn_analysis}). In this case, the BSIF classification accuracy was consistently higher, presenting also a smaller variability in the results than the Raw input image. These results suggest a better capability of BSIF filters to capture discriminating features of attack images. Along with the BSIF scale/depth analysis, results  indicate there is a potential gain when using some BSIF filters, and this  encouraged us to develop our fusion strategy explained ahead.

\subsection{Classification using lightweight CNNs}
\label{sec:our_cnn}

\begin{figure*}[!htb]
    \centering
    \includegraphics[width=\linewidth,
                     trim=0.9cm 0.7cm 0.5cm 0.5cm,clip]{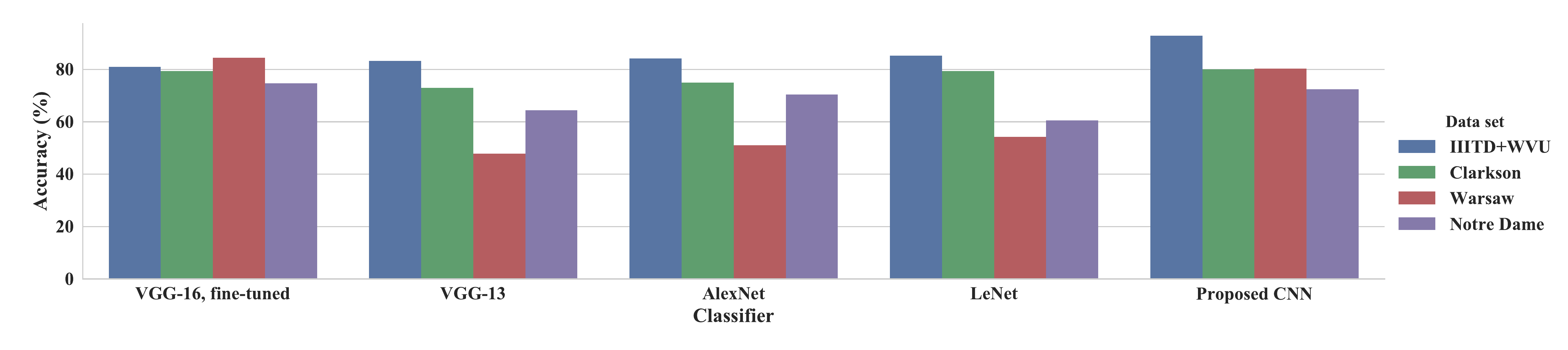}
    \caption{Performance of CNN architectures on the \emph{test unknown} partition, fine-tuned or trained from scratch in the \emph{train} partition of each dataset. Our proposed architecture has fewer layers, takes less time to train, and results in similar or superior accuracy to other known CNN architectures.}
    \label{fig:other_cnns}
\end{figure*}

As already mentioned, we train a CNN model for each of 60 BSIF transformations of the original image and a CNN for the original image, for a total of 61 CNNs.
As we will show in our experiments, the ensemble of the most complementary CNN models allows us to exploit different aspects of the iris image texture, which is important to achieve high detection rates, especially in challenging scenarios, such as cross-sensor and cross-dataset evaluation protocols.


Given that we do not have large training datasets, and also to avoid possible overfitting to specific datasets through the use of millions of parameters, we opt to use a not-too-deep network architecture. Experiments with deeper configurations did not translate into improved results in our case (Fig. \ref{fig:other_cnns}). Inspired by Menotti~\etal~\cite{Menotti:TIFS:2015}, our CNN architecture comprises two convolutional layers, and two fully-connected layers. Batch normalization~\cite{ioffe2015batch} is applied after each convolutional layer, to optimize the training procedure. The input of the network is an image of configurable size. The first convolutional layer consists of 16 filters of size $3 \times 3$ and Rectified Linear Unit (\emph{ReLU}) activation. Next, MaxPooling on a $9 \times 9$ pixel window and stride 2 is applied, just before the first batch normalization. The second convolutional layer comprises 32 filters of size $3 \times 3$ and \emph{ReLU} activation. 
MaxPooling on a $9 \times 9$ area with stride 8 and batch normalization follow this layer. The output of convolutional layers is fed into a fully-connected layer composed of 1,024 \emph{ReLU} activated neurons. Finally, the output layer is formed by two units, corresponding to two classes we want to recognize (authentic iris vs. presentation attack iris) following the SoftMax transformation. 

\subsection{Ensemble Fusion of Multiple Views}

In this section, we present the proposed approaches for fusing the result of the ensemble of CNNs.

\subsubsection{Random Forest Fusion}

Random Forests is a well-known ensemble method that has been used for classifier fusion~\cite{kuncheva2014} as well as for variable importance ranking~\cite{breiman2001random, genuer2010variable}. 
It is composed of multiple decision trees built on different random subsets of the training data.
Some of these decision trees tend to grow deep, sometimes leading to overfitting. This effect can be reduced by balancing the output of the decision trees and through tree pruning. By using a random forest classifier on top of the results of the different multi-view-CNN predictors, we can consolidate their outputs into a single prediction measure. As byproduct of this fusion, the random forest is also able to provide a ranking of predictor's importance, which can help us to select the most relevant predictors for our task.

Random Forests can estimate the test error (generalization error) of the ensemble, without needing to keep a separate test data partition. This is called out-of-bag (OOB) error estimation: the prediction error is calculated on all the samples that are left out of the bootstrap for each of the decision trees. 
During the process of training a Random Forest, after each tree is constructed, OOB error rate (of all variables) is compared to the classification error of the permutation of out-of-bag examples for each of the variables. As a result, we get an estimate of how much the misclassification error increases if each of the variables is disturbed. This measure is called variable importance.

\subsubsection{Voting Fusion}

Another strategy we exploit for decision-making is fusion through voting. We considered \emph{majority} and \emph{weighted voting}. 
In its simplest form, 
majority voting~\cite{kuncheva2014, kittler1998combining} considers all 61 CNNs as inputs to decide whether or not a given sample should be considered as an authentic or an attack. The Condorcet Jury Theorem states that if all classifiers produce independent predictions, and if each has a probability of correct prediction that is greater than 0.5, the addition of more voters will increase the probability that the consensus will make the correct decision \cite{kuncheva2014}. Veering away from simple majority voting, in some cases it is desirable to give greater weights to classifiers that are more likely to yield the right decision. With this in mind, we also use a strategy called~\emph{Best-Worst Weighted Vote} (\textit{BWWV})~\cite{Moreno-Seco2006}. The best and the worst ranking predictors are identified and receive maximum and minimum weight (1 and 0, respectively). All remaining predictor weights are linearly cast between these extremes.

We consider two ways to attribute weights to CNN predictors: classification accuracy and rank importance. For the former, predictors are sorted by decreasing accuracy for the weight assignment. For the latter, predictors are sorted by their rank importance calculated through the Random Forest classifier. We refer to these techniques as \texttt{BWWVA} and \texttt{BWWVI}, respectively.

\subsubsection{Classifier selection and Meta-fusion}
\label{subsec:meta_fusion_algorithm}

One would naturally expect that some predictors provide complementary views on the problem while other predictors are highly correlated.
In this section, we present our proposed algorithm for selecting the most relevant subset of predictors to use as the final ensemble for the PAD. We take advantage of {Gini} importance~\cite{Breiman:1984}, measured with tree-based models, and inter-rater agreement, measured using Cohen's kappa statistic~\cite{Fleiss:EPM:1973}.

\begin{figure}[t]
\centering
    \includegraphics[width=0.99\linewidth]{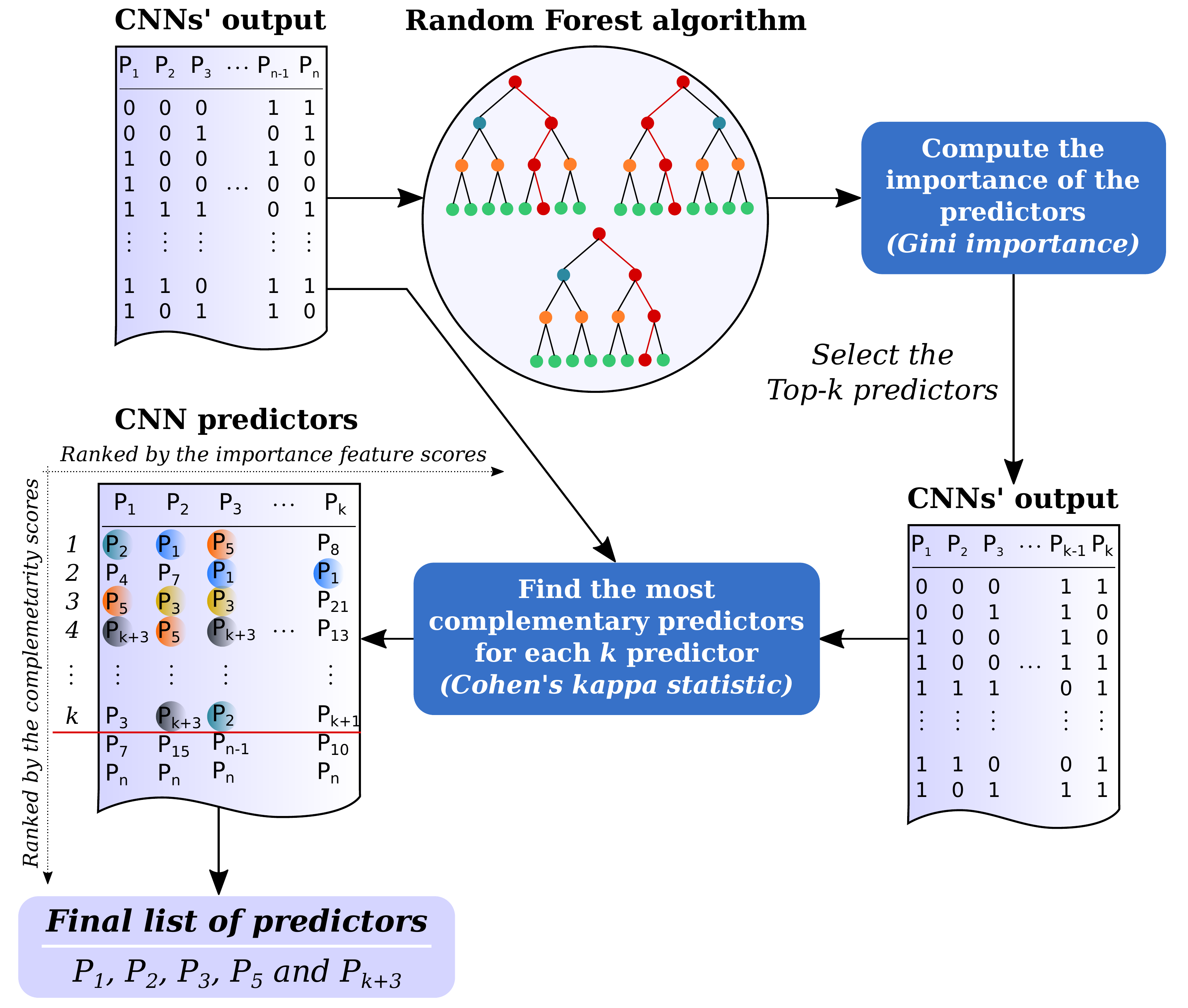}
    \caption{Proposed algorithm for selecting predictors: after generating CNN outputs, a Random Forest is trained on the binary results of all predictors (feature vector $v \in R^{61}$ dimensions), to rank predictors by importance. The most complementary to each of the $k$ most important predictors are calculated through Cohen's kappa statistic. Finally, a predictor list is generated by selecting those which appear in two or more columns in the calculated matrix.}
\label{fig:proposed_method_selection}
\end{figure}

Gini importance should give us the most promising predictors in terms of classification accuracy, while the inter-rater agreement measure should give us the most complementary predictors. Fig.~\ref{fig:proposed_method_selection} illustrates the proposed approach.

Our algorithm performs a score analysis of all multi-view-CNN predictors' outputs using the Random Forest algorithm to find the $k$ most important predictors. This step focuses on which models are important in the classification task in terms of their {Gini} importance, also known as mean decrease in impurity (MDI)~\cite{Breiman:1984}. In the context of our meta-analysis, each node of a tree represents a single feature, which is the output of a given predictor.

{Gini} importance measures how much the predictors decrease the weighted impurity in a tree. More precisely, the RF algorithm is a collection of decision trees aiming at splitting the data into two branches so that similar patterns end up in the same branch. The RF algorithm computes, during the training process, how much each predictor decreases the weighted impurity in a tree, and this metric is used to measure the quality of such splits. Our proposed algorithm takes advantage of that by ranking the multi-view-CNN predictors according to this measure and selecting the top-\textit{k} predictors, which will be our initial pool of good candidates for the fusion step.

After finding the $k$ most important models, we compute, through the Cohen's kappa statistic, which gives an agreement estimation between two predictors, their $l$ most complementary predictors, ending up with a $k \times l$ matrix $\mathbf{C}$ of predictors. 
We compute a final list of predictors by selecting those that appear in two or more columns of $\mathbf{C}$. The final decision-making is accomplished through SVM meta-fusion, which takes $k$ selected predictor outcomes as the input.
\section{Experimental Results}
\label{sec:experimental_results}

In this section, we describe the datasets, validation protocols and experimental results. 

\begin{table*}[!htbp]
\centering
\caption{Composition of the Datasets}
\label{tab:dataset_composition}
\resizebox{\linewidth}{!}{%
    \begin{tabular}{l|ccc|ccc|ccc}
    \multirow{2}{*}{{\bf Dataset}} & \multicolumn{3}{c|}{\bf Train} & \multicolumn{3}{c|}{\bf Test known} & \multicolumn{3}{c}{\bf Test unknown} \\
     & {\bf Live} & {\bf Contacts} & {\bf Printouts} & {\bf Live} & {\bf Contacts} & {\bf Printouts} & {\bf Live} & {\bf Contacts} & {\bf Printouts} \\ \hline\hline
    Clarkson & 2,469 & 1,122 & 1,346 & 1,485 & 765 & 908 & 638 & 494 & 144 \\ \hline
    IIITD-WVU & 2,250 & 1,000 & 3,000 & -- & -- & -- & 702 & 701 & 2,806 \\ \hline
    Notre Dame & 600 & 600 & -- & 900 & 900 & -- & 900 & 900 & -- \\ \hline
    Warsaw & 1,844 & -- & 2,669 & 974 & -- & 2,016 & 2,350 & -- & 2,160 \\ \hline\hline
    Combined & 7,163 & 2,722 & 7,015 & 3,359 & 1,665 & 2,924 & 4,590 & 2,095 & 5,110 \\
    \end{tabular}%
    }
\end{table*}

\subsection{Datasets}
\label{sec:dataset}

Our work was performed on datasets made available in the Iris Liveness Detection Competition 2017 \cite{Yambay2017}. There is one set from each of four universities involved --- Clarkson University, Warsaw University of Technology, IIITD-WVU and University of Notre Dame.

The Clarkson dataset \cite{Yambay2014, Yambay2015, Yambay2017} contains images of live irises, textured contact lenses, and iris printouts. The Warsaw dataset \cite{Yambay2017} comprises images of live irises and iris printouts. 
The Notre Dame dataset \cite{Doyle2015} contains images of live irises without textured contact lenses and of irises wearing textured contact lenses. Finally, the IIITD-WVU dataset \cite{Kohli2013, Gupta2014, Yadav2014, Kohli2016} contains images of live irises, textured contact lenses, iris printouts, and printouts of textured contact lenses.
Table \ref{tab:dataset_composition} summarizes the composition of the datasets. For some of our experiments, we also formed a merged dataset by combining the four sets, which we will refer to as the ``Combined'' dataset. The original cross-validation partitioning was kept the same in the ``Combined'' dataset.

\begin{figure*}[!htb]
    \centering
    \begin{subfigure}[b]{0.99\linewidth}
        \includegraphics[width=\linewidth, 
                         trim=0.4cm 0.5cm 0.8cm 0.3cm,clip]{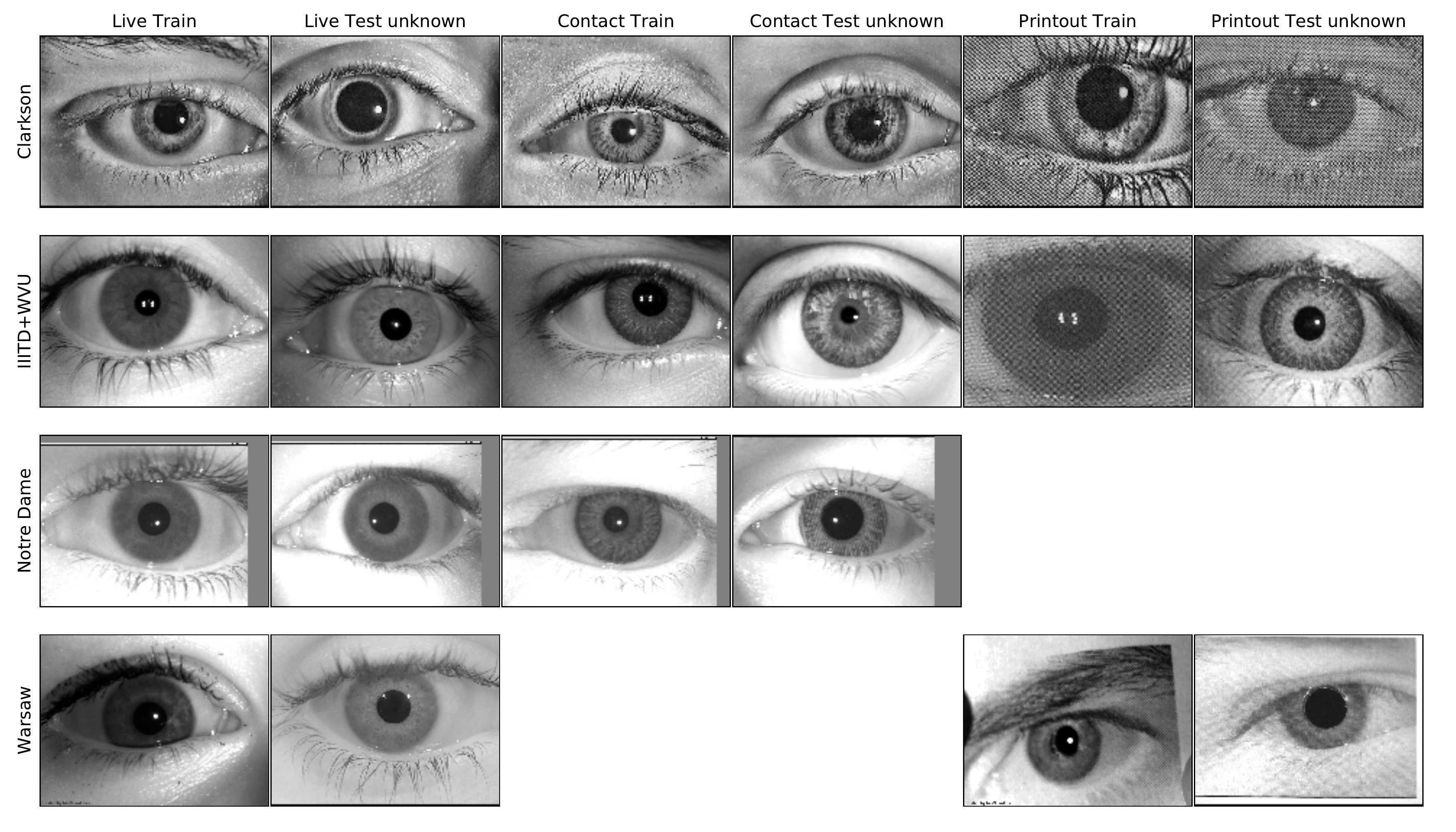}
    \end{subfigure}
    \caption{Image samples from all datasets, from the \emph{train} and \emph{unknown} partitions. The difference between train and unknown images, especially in the case of attacks, illustrates situations where classifiers commonly fail.}
    \label{fig:img_samples}
\end{figure*}

Each dataset is composed of a \emph{train} partition, made available to the participants to facilitate training their algorithms, and a \emph{test} partition, not distributed before the competition was ended, and used by the organizers to evaluate the submissions. LivDet-Iris 2017 co-organizers marked their test samples to form two groups of images. In the first group, referred to as \emph{test known}, both live images and images of artifacts had the same ``known'' properties as train samples. The images belonging to a second group, \emph{test unknown}, have different, or ``unknown'', properties than pictures included in the train subsets. Competition organizers applied different strategies when producing the \emph{test unknown} samples. Clarkson University included visible-light image printouts and new patterns of textured contact lenses, Warsaw University of Technology used different equipment to prepare and photograph iris printouts. The images of patterned contact lenses offered by University of Notre Dame are of different brands than those in the train set. Finally, the whole test partition of the IIITD-WVU benchmark is considered as \emph{test unknown}, since it was collected by a different institution (WVU) than the train set (IIITD), by a different sensor, and included outdoor acquisitions. Figure \ref{fig:img_samples} shows sample images of each dataset.

To keep our evaluation protocol compliant with LivDet-Iris 2017, training of our methods uses solely pre-defined training partitions of the datasets, while the final performance is estimated both on the \emph{test known} and \emph{test unknown} partitions.

\subsection{Experimental Protocol}
\label{sec:experimental_protocol}

Our experiments followed the LivDet-Iris 2017 competition protocol \cite{livdet2017}, using the same datasets and train/test partitions as described in Section \ref{sec:dataset}. A sub-partition consisting of a randomly selected 20\% of the images from each train set was created to serve as a validation set during training. The system is trained to perform binary classification: authentic iris images should be labeled as \emph{live}, while attack images (textured contact lenses, printouts of live images or printouts of textured contact lenses) should be labeled as \emph{attack}.

We measure the performance of classifiers using four metrics: 

\begin{itemize}
    \item \emph{Accuracy}, which is the ratio between the number of correctly classified images and the total number of images classified,
    \item \emph{Bona-Fide Presentation Classification Error Rate} (BPCER), which is the proportion of \emph{live} images that were incorrectly classified as \emph{attacks},  
    \item \emph{Attack Presentation Classification Error Rate} (APCER), which is the proportion of \emph{attack} images incorrectly classified as \emph{live} samples, and
    \item \emph{Half Total Error Rate} (HTER), which corresponds to the average of BPCER and APCER.
\end{itemize}

BPCER and APCER error rates were defined by ISO/IEC 30107-3 \cite{ISO_30107-3_2017} and adopted in LivDet-Iris 2017 for evaluation of submissions. The \emph{accuracy} and HTER are used in the training stage for ranking the obtained solutions.

The CNN classifiers output a liveness score in the range of $\langle 0,1 \rangle$ and the decision threshold was 0.5, as defined in the LivDet-Iris 2017 protocol. The source-code to reproduce all of our experiments is available to researchers \footnote{Available on GitHub: \url{https://github.com/akuehlka/mvlc_ipad}.}.

\subsection{Evaluation of BSIF Representations and CNN-based Detectors}
\label{sec:evaluation_bsif_cnns}


\begin{figure}[!hb]
    \centering
    \includegraphics[
        width=\linewidth,
        trim=0.6cm 0.7cm 0.5cm 0.5cm
        ]{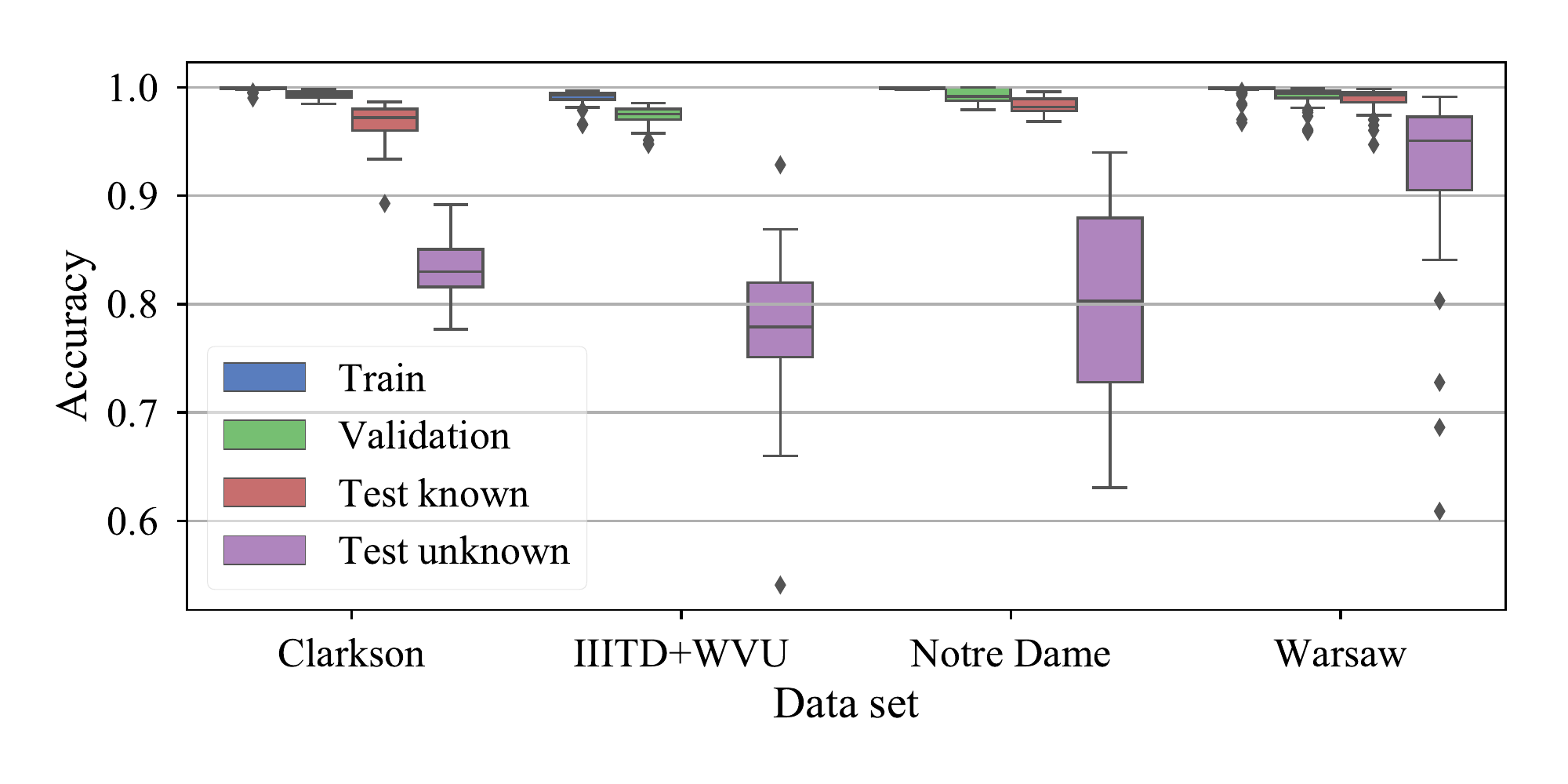}
    \caption{Distribution of classification accuracies for individual CNN predictors, on each dataset and partition.}
    \label{fig:individual_accuracy}
\end{figure}

The first part of our work was to train and evaluate 61 lightweight CNNs, which act as primary predictors in our system.
As it could be anticipated, individual CNN predictors result in a very good accuracy on the \emph{train} and \emph{test known} partitions, but they do not generalize well to \emph{test unknown} data. This can be seen in Fig.~\ref{fig:individual_accuracy}, which shows the distribution of accuracies obtained by all 61 CNN-based predictors.

Since the CNN is different for different feature descriptors, one can assume the variation in performance is caused by the ability of such descriptors to capture particular textures of the images. This may suggest that some filters are particularly good at classifying a certain type of attack (a certain textured lens brand or a type of a printout), but they are inadequate for others. The proposed method identifies good predictors and aggregates them in order to create a more robust cross-dataset PAD system.

Table ~\ref{tab:individual_top3} shows the top three CNN classifiers on each data set. The size of the BSIF filters that performed best in each data set may offer some insight about the types of features that are being used for classification. While CNN predictors achieved very high accuracy on Warsaw dataset using BSIF filters of size ranging from 5 to 7, the best performing filters on Notre Dame dataset are of size ranging from 3 to 5, while for Clarkson data the best filters are larger, and their size ranges from 7 to 15. The range of BSIF filter sizes that performed best in IIITD+WVU goes from 5 through 17, which is consistent with the wide diversity of images found in this dataset.

It is important to observe a basic difference in the datasets: while Warsaw attack images are printouts, Notre Dame's attack images are textured contact lenses. On the other hand, Clarkson and IIITD+WVU attack subsets are composed of a mix of textured contacts and printouts. Another fact that should be noted is that these data sets do not have similar proportions regarding the number of samples for each class in each partition. As an example, while there are only 144 printouts ($\sim11\%$) in Clarkson unknown, IIITD+WVU has 2,806 ($\sim66\%$) printouts in the same partition. These differences in data set composition may help to explain how different BSIF filter sizes achieve different ranges of performance for each dataset.


\begin{table}[!b]
\centering
\caption{Classification accuracy (\%) of the top 3 predictors for each dataset on the \textit{Test unknown} partition.}
\label{tab:individual_top3}
\begin{tabular}{cccc}
\hline
Dataset                     & BSIF Filter & Accuracy & HTER    \\ \hline \hline
\multirow{3}{*}{Clarkson}   & 15x15x11    & 87.08\%  & 12.91\% \\
                            & 07x07x10    & 82.69\%  & 17.29\% \\
                            & 13x13x12    & 80.74\%  & 19.25\% \\ \midrule
\multirow{3}{*}{IIITD+WVU}  & 05x05x07    & 81.87\%  & 47.45\% \\
                            & 09x09x11    & 78.69\%  & 27.03\% \\
                            & 17x17x07    & 77.62\%  & 19.41\% \\ \midrule
\multirow{3}{*}{Notre Dame} & 05x05x11    & 71.78\%  & 28.22\% \\
                            & 05x05x10    & 69.11\%  & 30.89\% \\
                            & 03x03x06    & 67.83\%  & 32.17\% \\ \midrule
\multirow{3}{*}{Warsaw}     & 05x05x11    & 98.00\%  & 2.08\% \\ 
                            & 07x07x12    & 93.84\%  & 6.24\% \\
                            & 05x05x08    & 91.95\%  & 8.26\% \\
\hline
\end{tabular}
\end{table}

\subsection{Fusion Evaluation}
\label{sec:eval_simple_fusion}

Before deciding that a more elaborate method for fusion would be required, we experimented with basic methods. Table \ref{tab:simple_fusion} summarizes results of four basic fusion methods: Random Forest (RF), Majority Voting (MV), Best-to-worst Weighted Voting by Accuracy (BWWVA), and by Importance (BWWVI).
While there is no clear trend towards a specific fusion method, all of them seem to perform best in specific datasets and partitions. With regard to the data sets, the same trend seen in the CNN predictors happens here: Warsaw had the highest accuracy, followed by IIITD+WVU, Clarkson, and Notre Dame. However, in some cases simple fusion led to an obvious gain in accuracy that was not always the case: some fusion methods were not able to obtain a better result than the best individual classifier, typically in the unknown test partitions.

\begin{table*}[!htb]
\centering
\caption{Results for four simple fusion strategies on the \emph{test known} (K) and \emph{test unknown} (U) partitions; best HTERs in bold.
}
\label{tab:simple_fusion}
\resizebox{\textwidth}{!}{%
\huge
\begin{tabular}{ll|cccc|cccc|cccc|cccc} \hline
                            &   & \multicolumn{4}{c|}{\textbf{RF}}           & \multicolumn{4}{c|}{\textbf{MV}}           & \multicolumn{4}{c|}{\textbf{BWWVA}}        & \multicolumn{4}{c}{\textbf{BWWVI}}        \\ \cline{3-18}
                            &   & Accuracy & APCER & BPCER & HTER  & Accuracy & APCER & BPCER & HTER  & Accuracy & APCER & BPCER & HTER  & Accuracy & APCER & BPCER & HTER  \\ \hline
\multirow{2}{*}{Clarkson}   & K & 97.74    & 4.45  & 0.74  & 2.59  & 99.48    & 0.00  & 0.87  & \textbf{0.44}  & 99.44    & 0.00  & 0.94  & 0.47  & 99.40    & 0.09  & 0.94  & 0.52  \\
                            & U & 78.70    & 41.94 & 0.63  & 21.28 & 85.59    & 28.17 & 0.63  & 14.40 & 86.37    & 26.60 & 0.63  & \textbf{13.61} & 85.75    & 27.70 & 0.78  & 14.24 \\ \hline
\multirow{2}{*}{IIITD+WVU}  & K & -        & -     & -     & -     & -        & -     & -     & -     & -        & -     & -     & -     & -        & -     & -     & -     \\
                            & U & 83.34    & 15.94 & 20.23 & \textbf{18.08} & 83.82    & 12.83 & 32.91 & 22.87 & 84.25    & 12.29 & 33.05 & 22.67 & 83.06    & 9.21  & 55.56 & 32.38 \\ \hline
\multirow{2}{*}{Notre Dame} & K & 99.44    & 0.44  & 0.67  & \textbf{0.56}  & 99.33    & 0.22  & 1.11  & 0.67  & 99.39    & 0.33  & 0.89  & 0.61  & 99.27    & 0.44  & 1.00  & 0.72  \\
                            & U & 85.11    & 29.33 & 0.44  & \textbf{14.89} & 82.39    & 34.22 & 1.00  & 17.61 & 81.61    & 35.78 & 1.00  & 18.39 & 82.28    & 34.44 & 1.00  & 17.72 \\ \hline
\multirow{2}{*}{Warsaw}     & K & 99.80    & 0.05  & 0.51  & 0.28  & 99.87    & 0.04  & 0.31  & 0.18  & 99.87    & 0.05  & 0.31  & 0.18  & 99.90    & 0.00  & 0.30  & \textbf{0.15}  \\
                            & U & 99.51    & 0.32  & 0.64  & \textbf{0.48}  & 99.49    & 0.37  & 0.64  & 0.50  & 99.51    & 0.37  & 0.59  & \textbf{0.48}  & 99.46    & 0.23  & 0.81  & 0.52  \\ \hline
\end{tabular}
}%
\end{table*}

RF fusion was the most successful method, outperforming the others in three situations. 
With the exception of BWWVA, which tied with RF in the unknown partition of Warsaw, all other fusion methods were best in a single dataset/partition.
\begin{figure*}[b]
    \centering
    \includegraphics[
        width=\textwidth,
        trim=0.5cm 0cm 0.5cm 1cm
    ]{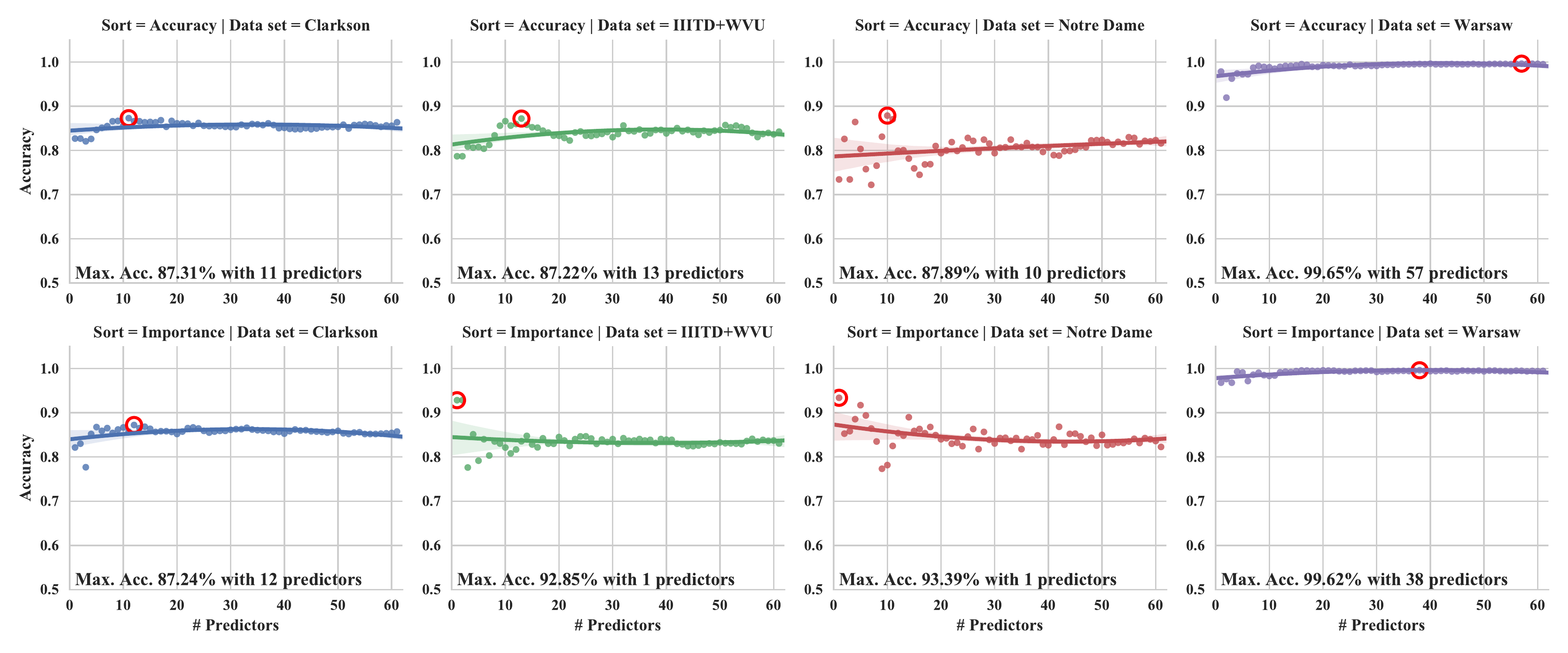}
    \caption{Relationship between classification accuracy and number of predictors used in the fusion. The top row shows results for \textit{BWWVA}, and the bottom row for \textit{BWWVI}.}
    \label{fig:incremental_fusion}
\end{figure*}

These somehow contradictory results motivated us to further investigate this aspect.  Fig.~\ref{fig:incremental_fusion} shows the relationship between the number of predictors and the output accuracy, for each weighted voting fusion technique and for each data set. 
For each fusion method and dataset, after the optimal number of predictors is reached, the addition of extra predictors does not improve the results, and in some cases it may even be detrimental.
In some cases, the optimal accuracy is reached using a single predictor, but other cases may require up to 57 predictors. 

A method for predictor selection is necessary so that we can obtain the best accuracy from the ensemble.
However, this process is not trivial: the analysis of the predictor-accuracy relation in other data partitions (train and validation) reveals significantly different trends. Consequently, it is not always possible to determine the optimal number of predictors to be used in the unknown partition by simply using information from other partitions.

Accuracy is the obvious metric of choice for selecting the best base classifiers, but it does not tell us about the complementarity relations between different classifiers. The \textit{Gini} importance measure calculated by training a Random Forest can suggest the best classifiers, based on how much each predictor helps to reduce the impurity in a decision tree. Therefore, it can also be used as a criterion for predictor selection. However, neither of these estimations helps us to determine an optimal number of predictors. Furthermore, our results show there is no distinct advantage in using either accuracy or importance. These results motivated us to search for a method for selection of predictors that is based on their already known properties, but also based on how different base classifiers integrate with each other.

\subsection{Cross-Domain Evaluation}
\label{sec:cross_dataset_evaluation}

With the exception of Clarkson, all LivDet-Iris 2017 datasets were designed to allow cross-sensor evaluation. Images in their unknown partitions were captured with different sensors and environments. Additionally, we conducted cross-dataset experiments to verify the possibility of transfer learning across these datasets. However, direct cross-dataset evaluation using simple fusion did not result in good accuracy. In fact, training CNN predictors in one data set and testing them in another resulted in accuracy no better than random prediction.

One of the premises of our approach is the use of multiple views of the data, in order to capture texture subtleties. We can confirm this if we consider the ranges of BSIF filter sizes that had better accuracy in the different data sets. This suggests there is enough difference in texture scale from one data set to another to cause the individual CNN predictors not to be able to recognize them. The variation of these data sets in nature (texture contacts, printouts, or a mix of both), composition (regarding sizes of partitions and classes), acquisition devices and environments can explain the limited ability for transfer learning here.

Despite the fact that direct cross-dataset evaluation was not successful, training CNN predictors in a combined dataset produced much better results. Table \ref{tab:combined_results} shows a comparison of all predictors and their fusion in the combined dataset. Both individual CNNs and simple fusion methods achieved typically over $95\%$ classification accuracy in the known test partition. In the unknown partition, while most CNN predictors achieved HTER of $12\%$ or lower, all simple fusion methods obtained HTER below $8\%$, with classification accuracies higher than $91\%$ even on the unknown set. 

\begin{table}[!ht]
\centering
\caption{HTER (\%) on the Combined dataset.}
\label{tab:combined_results}
\begin{tabular}{c|ccccc}
\hline
  &No Fusion & RF    & MV    & BWWVI & BWWVA \\
  &(avg.) \\ \hline
K & 2.57  & 0.74  & 0.65  & 0.59  & 0.63 \\
U & 12.47 & 7.02  & 6.89  & 7.12  & 6.88 \\ \hline
\end{tabular}
\end{table}


\begin{figure*}[ht]
    \centering
    \includegraphics[
        width=\linewidth,
        trim=0.5cm 1cm 0.5cm 0.5cm
    ]{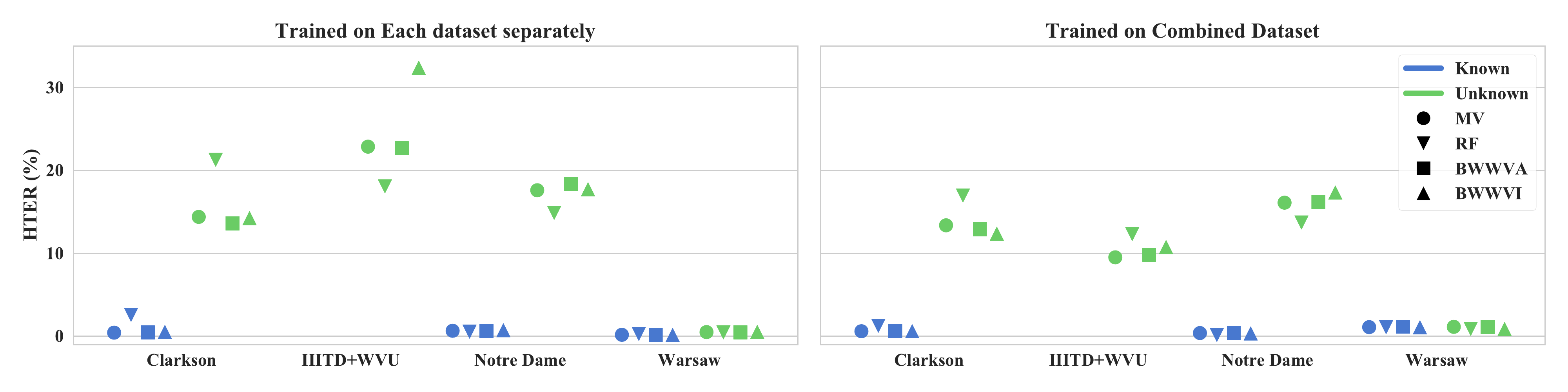}
    \caption{HTER distribution for fusion methods. On the left, training and evaluation is performed on each individual dataset. On the right, training is performed on the combined dataset, and evaluation is performed on each individual dataset.}
    \label{fig:fusion_distrib}
\end{figure*}

Fig.~\ref{fig:fusion_distrib} shows the distribution of HTER for fusion methods, in a comparison between training on each individual dataset, and training on the combined dataset. Performing the training on the combined dataset was particularly favorable in IIITD+VWU dataset, in which HTER was reduced from more than $25\%$ to nearly $10\%$. 

\subsection{Evaluation of the Proposed Meta-Analysis: Selection and Meta-Fusion of Classifiers}
\label{sec:evaluation_meta_analysis}

\begin{table*}[!hb]
\centering
\caption{Performance results (\%) of the proposed method, majority vote, and the best individual CNN model for the \emph{test known} (K), \emph{test unknown} (U), and \emph{overall test} (O) partitions from datasets used in this work. The classifiers were trained on the train partition of each dataset. $^\dagger$~IIITD-WVU dataset contains only the \emph{unknown test} partition.}
\setlength{\tabcolsep}{3.5pt}
\begin{tabular}{cc||c|c|c||c|c|c||c|c|c||c|c|c}
    \toprule
				     & 						& \multicolumn{3}{c||}{\textbf{Meta-Fusion via SVM}} 
										    & \multicolumn{3}{c||}{\textbf{BWWVA}}
										    & \multicolumn{3}{c||}{\textbf{RF}}
										    & \multicolumn{3}{c}{\textbf{Best Individual CNN}} \\

    \textbf{Dataset} & \textbf{Testing set} & \textbf{APCER} & \textbf{BPCER} & \textbf{HTER} 
                                            & \textbf{APCER} & \textbf{BPCER} & \textbf{HTER} 
                                            & \textbf{APCER} & \textbf{BPCER} & \textbf{HTER} 
                                            & \textbf{APCER} & \textbf{BPCER} & \textbf{HTER} \\
    \hline\hline
			                    & K    & 0.00    & 2.44    & 1.22    & 0.33    & 0.89    & 0.61	   & 0.44   & 0.67  & 0.56  & 0.67    & 2.44    & 1.56    \\
    Notre Dame            		& U    & 9.22    & 1.44    & 5.33    & 35.78   & 1.00    & 18.39   & 29.33  & 0.44  & 14.89 & 32.44   & 2.44    & 17.44   \\
							    & O    & 4.61    & 1.94    & 3.28    & 18.06   & 0.94    & 9.50	   & 14.89  & 0.56  & 7.72  & 16.56   & 2.44    & 9.50    \\
    \midrule                               	
								& K    & 0.05    & 0.82    & 0.44    & 0.05    & 0.31	 & 0.18	   & 0.05   & 0.51  & 0.28  & 0.30    & 0.41    & 0.35    \\
    Warsaw                		& U    & 0.09    & 1.49    & 0.79    & 0.37	   & 0.60	 & 0.48    & 0.32   & 0.64  & 0.48  & 20.97   & 0.68    & 10.83   \\
						        & O    & 0.07    & 1.29    & 0.68    & 0.21	   & 0.45	 & 0.33	   & 0.19   & 0.58  & 0.38  & 10.63   & 0.55    & 5.59    \\
    \midrule                    
                                & K    & 4.55    & 0.20    & 2.37    & 0.00    & 0.94    & 0.47	   & 3.48   & 0.47  & 1.98  & 1.26    & 1.75    & 1.50    \\
    Clarkson              		& U    & 41.54   & 0.31    & 20.92   & 26.30   & 0.63    & 13.47   & 39.12  & 0.63  & 19.88 & 32.71   & 1.88    & 17.29   \\
								& O    & 18.66   & 0.24    & 9.45	 & 13.30   & 0.78    & 7.04	   & 21.30  & 0.55  & 10.93 & 16.98   & 1.82    & 9.40    \\
    \midrule
								& U    &  	     &         &         &         &         &         &        &       &       &         &         &         \\
    \multirow{-2}{*}{IIITD+WVU~$^\dagger$} 
                                & O    & \multirow{-2}{*}{12.32}	
												 & \multirow{-2}{*}{17.52}
												 & \multirow{-2}{*}{14.92}
												 & \multirow{-2}{*}{12.29}
												 & \multirow{-2}{*}{33.05}
												 & \multirow{-2}{*}{22.67}
												 & \multirow{-2}{*}{5.56}
												 & \multirow{-2}{*}{21.23}
												 & \multirow{-2}{*}{13.39}
												 & \multirow{-2}{*}{21.81}
												 & \multirow{-2}{*}{72.22}
												 & \multirow{-2}{*}{47.02} \\
    \hline
    \hline
    Average						& O    &\bf 8.92 &\bf 5.25 &\bf 7.08 & 10.96   & 8.80   & 9.88   & 10.48    & 5.73  & 8.11  & 16.50	& 19.26	  & 17.88     \\
    \bottomrule
\end{tabular}\\
\label{tab:meta_classification_results}
\end{table*}

In this section, we evaluate the proposed method designed to automatically select the most relevant multi-view-CNN predictors, in terms of their importance and complementarity, and to fuse their individual results via meta-fusion approach, which was performed using the Support Vector Machine.

First, we used our proposed algorithm described in Section~\ref{subsec:meta_fusion_algorithm} to select the most relevant predictors from a pool of $61$ predictors. Next, we performed a meta-fusion using the SVM with a radial basis function kernel. Both the parameters of the SVM and the parameter \textit{k} (see Sec.~\ref{subsec:meta_fusion_algorithm}) were found through grid search in the aggregated training sets of all datasets considered in this work. The \emph{known} and \emph{unknown} test sets were used only to report the final performance results.

\mathchardef\mhyphen="2D  
We performed a fine grid-search of parameter \textit{k} on the neighborhood of (5, 20), and the best HTER was achieved with $k=16$.
According to one-sample Wilcoxon signed-rank test, the observed differences in the HTER values obtained for each $k$ are statistically significant at the significance level $\alpha=0.05$ ($p\mhyphen value=0.0004$). Henceforth, we use $k=16$ to report performance results of our proposed method.

Furthermore, Table~\ref{tab:meta_classification_results} shows the effectiveness of our meta-fusion approach compared to the best multi-view-CNN predictor and the majority vote fusion technique. In addition to the results for \emph{test known} and \emph{test unknown} partitions, Table \ref{tab:meta_classification_results} also presents \emph{overall test}, which corresponds to the accuracy obtained on both test partitions combined. These results are in agreement with those reported in the literature~\cite{kuncheva2014}, which states that both accuracy and complementarity are the foundation for effective fusion.

\subsection{Comparison with the State of the Art}
\label{sec:comparison_sota}

In this section, we compare our results with the LivDet-Iris 2017 best performing method. As we can see, our method outperforms the competition winner. Table \ref{tab:livdet} summarizes the results of the competition in contrast to our best solution developed solely on the LivDet-Iris 2017 train partitions.

Even in the cases where our methods were not able to outperform one of the specific error rates, we were able to significantly reduce the combined error rate. This happens, for instance, in the overall performance on IIITD+WVU dataset: the best HTER among LivDet-Iris 2017 competitors is $16.7\%$. Although our methods did not outperform their BPCER of $3.99\%$, we managed to lower HTER to $7.9\%$ on that specific dataset. Similar situations occurred on all datasets, showing a consistent improvement of our results over the state of the art.

\newcommand{\specialcell}[2][c]{%
  \begin{tabular}[#1]{@{}c@{}}#2\end{tabular}
}

\begin{table}[!htb]
	\centering
	\caption{Comparison of HTER (\%) with LivDet-Iris 2017, on the combined \emph{test known} and \emph{test unknown} partitions.}
	\label{tab:livdet}
	\begin{tabular}{cccc}
		\topline
						 &                           & \textbf{Proposed}      & \\
		\textbf{Dataset} & \textbf{LivDet-Iris 2017} & \textbf{Meta-Fusion}   & \textbf{Error} \\
						 & \textbf{Winner}           & \textbf{approach} 	  & \textbf{Reduction (\%)} \\
		\hline\hline
			Clarkson   & 9.59   & 9.45	&  1.46  \\
			\hline
			IIITD+WVU  & 16.70  & 14.92	&  10.66 \\
			\hline
			Notre Dame & 4.03   & 3.28	&  18.61  \\
			\hline
			Warsaw     & 5.81   & 0.68	&  88.30  \\
			\hline\hline
			Average    & 9.03   & 7.08	&  21.59   \\
			\bottomrule
	\end{tabular}
\end{table}

Meta-Fusion results were further improved when applied to the combined dataset. Table \ref{tab:combined-meta} presents these results in comparison with LivDet-Iris 2017, and also with our previous Meta-Fusion results. The overall HTER was reduced from $7\%$ to $4\%$ when Meta-Fusion was applied to the combined dataset. At this point, it is necessary to be clear about the LivDet-Iris 2017 comparison: the known and unknown HTER numbers presented in Table \ref{tab:combined-meta} are estimated. Since BPCER is not available for all \emph{known} and \emph{unknown} test partitions in LivDet-Iris 2017, we assumed 0 (perfect score) in our estimation. Even with this optimistic assumption about the competition results, our method achieved an error reduction of more than $50\%$ with regard to the former.

Our current implementation takes, on average, 0.028s to perform classification on a single image. Timing was performed on a 4-core Intel(R) Xeon(R) CPU E5-2650 v4 @ 2.20GHz with 128GB of RAM, equipped with a GeForce GTX 1080 Ti GPU. This time includes the CNN classification of the 61 data views and the meta-fusion of their results. Since the LivDet-Iris protocol does not establish parameters for speed efficiency analysis, the main directive for our implementation herein was classification accuracy. Therefore our method's efficiency can certainly be optimized, if needed. 

\begin{table}[ht]
    \begin{minipage}{\linewidth}
	\centering
	\caption{Results in terms of HTER (\%) for our Meta-Fusion approach trained on the Combined dataset. Error reduction (ER) values present the error decrease achieved by our approach with regard to LivDet-Iris 2017 winner.}
	\label{tab:combined-meta}
    \begin{tabular}{cccccc}
        \topline
                         &                           & \multicolumn{4}{c}{\textbf{Proposed Meta-Fusion approach (\%)}} \\
                         
        \cline{3-6}
                         & \textbf{LivDet-Iris 2017} & \multicolumn{2}{c}{\textbf{Trained on}} & \multicolumn{2}{c}{\textbf{Trained on}} \\
        \textbf{Dataset} & \textbf{Winner}           & \multicolumn{2}{c}{\textbf{each dataset}} & \multicolumn{2}{c}{\textbf{Combined dataset}} \\
						 & \textbf{HTER}             & \textbf{HTER} 	  & \textbf{ER}  & \textbf{HTER} 	  & \textbf{ER} \\
        \hline\hline
        K   & 0.74\footnote{Estimated\label{fn:1}}  & 1.34          & -81.08        & 0.74 & 0.00 \\
        U   & 13.23\textsuperscript{\ref{fn:1}}     & 10.49         & 20.71         & 8.39 & 36.58 \\
        O   & 9.03                                  & 7.08          & 21.59         & 4.44 & 50.83 \\
        \bottomrule
    \end{tabular}
    \end{minipage}
\end{table}

\section{Conclusions}
\label{sec:conclusions}

In this paper, we addressed the iris Presentation Attack Detection problem, as defined by the LivDet-Iris 2017 Competition, using an approach based on an ensemble of multi-view learning detectors. Our method has advanced the state of the art in iris PAD and offered insight on the potential use of different BSIF filters to deal with different textures in the same domain.

We presented a new approach combining two techniques for iris PAD: CNNs and Ensemble Learning. Extensive experimentation was conducted using the most challenging datasets publicly available. The experiments included cross-sensor and cross-dataset evaluations. Results show a varying ability for different BSIF+CNN representations to capture different aspects of the input images.

Simple fusion experiments show that although helpful, such techniques are not yet capable to provide optimal classification accuracy.  In fact, we demonstrate how the continuous addition of classifiers to the fusion does not necessarily improve the classification performance. In that context, we presented a new method for selection of classifiers, based on the meta-analysis of their Gini importance and inter-classifier complementarity.

Our Meta-Fusion method was able to consistently outperform the LivDet-Iris 2017 competition winner, with an overall Reduction Error Rate of more than $21\%$. Specifically, the HTER in the Warsaw dataset was only $0.68\%$, corresponding to a reduction in error of more than 88\% in relation to the top result reported in LivDet-Iris 2017. Although not as extreme as in Warsaw case, classification accuracy was also improved for other datasets, with reduction in error ranging from $1$ to $19\%$. Experiments with the combined dataset showed an additional improvement of $37\%$ on the overall HTER.

As a suggestion for future work, an immediate alternative application for our method is face PAD. A significant portion of face recognition attacks are based on printouts, as we believe meta-fusion of multi-view-CNN predictors could be easily applied to it with good potential for success.


\section{Acknowledgement}
\label{sec:ack}
This research was partially supported by the Brazilian Coordination for the Improvement of Higher Education Personnel (CAPES) through grants BEX 12976/13-0 and DeepEyes as well as by the S\~{a}o Paulo Research Foundation (FAPESP) through the D\'{e}j\`{a}Vu  research project (Grant \#2017/12646-3).

\bibliographystyle{IEEEtran}
\bibliography{iris-pad}

\begin{thebibliography}{10}
\providecommand{\url}[1]{#1}
\csname url@samestyle\endcsname
\providecommand{\newblock}{\relax}
\providecommand{\bibinfo}[2]{#2}
\providecommand{\BIBentrySTDinterwordspacing}{\spaceskip=0pt\relax}
\providecommand{\BIBentryALTinterwordstretchfactor}{4}
\providecommand{\BIBentryALTinterwordspacing}{\spaceskip=\fontdimen2\font plus
\BIBentryALTinterwordstretchfactor\fontdimen3\font minus
  \fontdimen4\font\relax}
\providecommand{\BIBforeignlanguage}[2]{{%
\expandafter\ifx\csname l@#1\endcsname\relax
\typeout{** WARNING: IEEEtran.bst: No hyphenation pattern has been}%
\typeout{** loaded for the language `#1'. Using the pattern for}%
\typeout{** the default language instead.}%
\else
\language=\csname l@#1\endcsname
\fi
#2}}
\providecommand{\BIBdecl}{\relax}
\BIBdecl

\bibitem{Daugman_PRS_2001}
J.~Daugman and C.~Downing, ``Epigenetic randomness, complexity and singularity
  of human iris patterns,'' \emph{Proceedings of the Royal Society B:
  Biological Sciences}, vol. 268, no. 1477, pp. 1737--1740, 2001.

\bibitem{Bowyer_BTAS_2016}
K.~W. Bowyer and P.~J. Flynn, ``Biometric identification of identical twins: A
  survey,'' in \emph{IEEE Intl. Conference on Biometrics Theory, Applications
  and Systems (BTAS)}, Sept 2016, pp. 1--8.

\bibitem{UIDAI}
{Unique Identification Authority of India, Government of India}, ``{Unique
  Identification Authority of India Website},'' \url{https://uidai.gov.in/},
  2018, [Online; accessed 02/12/2018].

\bibitem{Daugman:SPIE:2014}
J.~Daugman, ``{600 million citizens of India are now enrolled with biometric
  ID},'' in \emph{SPIE Newsroom}, 2014, pp. 1--4.

\bibitem{NEXUS_URL}
{Canada Border Services Agency and U.S. Customs and Border Protection},
  ``{NEXUS},'' \url{https://www.cbsa-asfc.gc.ca/prog/nexus/menu-eng.html},
  accessed January 19, 2018.

\bibitem{ISO_30107-1_2016}
{ISO/IEC 30107-1:2016}, ``{Information technology -- Biometric presentation
  attack detection -- Part 1: Framework}.''

\bibitem{Yambay2017}
D.~A. Yambay, B.~Becker, N.~Kohli, D.~Yadav, A.~Czajka, K.~W. Bowyer,
  S.~Schuckers, R.~Singh, M.~Vatsa, A.~Noore, D.~Gragnaniello, C.~Sansone,
  L.~Verdoliva, L.~He, Y.~Ru, H.~Li, N.~Liu, Z.~Sun, and T.~Tan, ``{LivDet 2017
  - Iris Liveness Detection Competition 2017},'' \emph{Biometrics: Theory
  Applications and Systems (BTAS)}, pp. 0--5, 2017.

\bibitem{Kannala_ICPR_2012}
J.~Kannala and E.~Rahtu, ``Bsif: Binarized statistical image features,'' in
  \emph{Intl. Conference on Pattern Recognition (ICPR)}, Nov 2012, pp.
  1363--1366.

\bibitem{Daugman_IMAIP_2000}
J.~Daugman, ``Wavelet demodulation codes, statistical independence, and pattern
  recognition,'' in \emph{Institute of Mathematics and its Applications,
  (IMA-IP)}, 2000, pp. 244--260.

\bibitem{Thalheim_CT_2002}
L.~Thalheim, J.~Krissler, and P.-M. Ziegler, ``{Biometric Access Protection
  Devices and their Programs Put to the Test, Available online in c't Magazine,
  No. 11/2002, p. 114},'' on-line, 2002.

\bibitem{Al-Raisi_TI_2008}
\BIBentryALTinterwordspacing
A.~N. Al-Raisi and A.~M. Al-Khouri, ``Iris recognition and the challenge of
  homeland and border control security in uae,'' \emph{Telematics and
  Informatics}, vol.~25, no.~2, pp. 117 -- 132, 2008. [Online]. Available:
  \url{http://www.sciencedirect.com/science/article/pii/S0736585306000360}
\BIBentrySTDinterwordspacing

\bibitem{Doyle_ICB_2013}
\BIBentryALTinterwordspacing
J.~S. Doyle, P.~J. Flynn, and K.~W. Bowyer, ``Automated classification of
  contact lens type in iris images,'' in \emph{{IEEE} Int. Conference on
  Biometrics (ICB)}, June 2013, pp. 1--6. [Online]. Available:
  \url{https://doi.org/10.1109/ICB.2013.6612954}
\BIBentrySTDinterwordspacing

\bibitem{HeXiaofu_ICB_2009}
\BIBentryALTinterwordspacing
X.~He, Y.~Lu, and P.~Shi, ``A new fake iris detection method,'' in \emph{{IEEE}
  Int. Conference on Biometrics (ICB)}, M.~Tistarelli and M.~S. Nixon,
  Eds.\hskip 1em plus 0.5em minus 0.4em\relax Berlin, Heidelberg: Springer
  Berlin Heidelberg, 2009, pp. 1132--1139. [Online]. Available:
  \url{http://dx.doi.org/10.1007/978-3-642-01793-3_114}
\BIBentrySTDinterwordspacing

\bibitem{Zuo_TIFS_2007}
J.~Zuo, N.~A. Schmid, and X.~Chen, ``On generation and analysis of synthetic
  iris images,'' \emph{{IEEE} Trans. Inf. Forens. Security}, vol.~2, no.~1, pp.
  77--90, March 2007.

\bibitem{Trokielewicz_BTAS_2016}
M.~Trokielewicz, A.~Czajka, and P.~Maciejewicz, ``Human iris recognition in
  post-mortem subjects: Study and database,'' in \emph{{IEEE} Int. Conference
  on Biometrics: Theory Applications and Systems (BTAS)}, Sept 2016, pp. 1--6.

\bibitem{Komulainen_IJCB_2014}
J.~Komulainen, A.~Hadid, and M.~Pietikäinen, ``Generalized textured contact
  lens detection by extracting bsif description from cartesian iris images,''
  in \emph{{IEEE} Int. Joint Conference on Biometrics (IJCB)}, Sept 2014, pp.
  1--7.

\bibitem{Lovish_CAIP_2015}
Lovish, A.~Nigam, B.~Kumar, and P.~Gupta, ``Robust contact lens detection using
  local phase quantization and binary gabor pattern,'' in \emph{Int. Conference
  on Computer Analysis of Images and Patterns (CAIP)}, G.~Azzopardi and
  N.~Petkov, Eds.\hskip 1em plus 0.5em minus 0.4em\relax Springer International
  Publishing, 2015, pp. 702--714.

\bibitem{Gragnaniello_TIFS_2015}
D.~Gragnaniello, G.~Poggi, C.~Sansone, and L.~Verdoliva, ``An investigation of
  local descriptors for biometric spoofing detection,'' \emph{{IEEE} Trans.
  Inf. Forens. Security}, vol.~10, no.~4, pp. 849--863, Apr. 2015.

\bibitem{Sequeira_TSP_2016}
A.~F. Sequeira, S.~Thavalengal, J.~Ferryman, P.~Corcoran, and J.~S. Cardoso,
  ``A realistic evaluation of iris presentation attack detection,'' in
  \emph{Int. Conference on Telecommunications and Signal Processing (TSP)},
  June 2016, pp. 660--664.

\bibitem{Gragnaniello_SITIS_2014}
D.~Gragnaniello, G.~Poggi, C.~Sansone, and L.~Verdoliva, ``Contact lens
  detection and classification in iris images through scale invariant
  descriptor,'' in \emph{Int. Conference on Signal-Image Technology
  Internet-Based Systems (SITIS)}, Nov 2014, pp. 560--565.

\bibitem{Pala_CVPR_2017}
F.~Pala and B.~Bhanu, ``Iris liveness detection by relative distance
  comparisons,'' in \emph{IEEE Conference on Computer Vision and Pattern
  Recognition (CVPR) Workshops}, July 2017.

\bibitem{Akhtar_AVSS_2014}
Z.~Akhtar, C.~Micheloni, C.~Piciarelli, and G.~L. Foresti, ``Mobio\_livdet:
  Mobile biometric liveness detection,'' in \emph{{IEEE} Int. Conference on
  Advanced Video and Signal Based Surveillance (AVSS)}, Aug 2014, pp. 187--192.

\bibitem{Chen_PRL_2012}
\BIBentryALTinterwordspacing
R.~Chen, X.~Lin, and T.~Ding, ``Liveness detection for iris recognition using
  multispectral images,'' \emph{Pattern Recognition Letters}, vol.~33, no.~12,
  pp. 1513 -- 1519, 2012. [Online]. Available:
  \url{http://www.sciencedirect.com/science/article/pii/S0167865512001262}
\BIBentrySTDinterwordspacing

\bibitem{Galbally_Handbook_2016}
\BIBentryALTinterwordspacing
J.~Galbally, M.~Savvides, S.~Venugopalan, and A.~A. Ross, \emph{Iris Image
  Reconstruction from Binary Templates}.\hskip 1em plus 0.5em minus 0.4em\relax
  London: Springer London, 2016, pp. 469--496. [Online]. Available:
  \url{http://dx.doi.org/10.1007/978-1-4471-6784-6_20}
\BIBentrySTDinterwordspacing

\bibitem{Silva_SIBGRAPI_2015}
P.~Silva, E.~Luz, R.~Baeta, H.~Pedrini, A.~X. Falc{\~{a}}o, and D.~Menotti,
  ``An approach to iris contact lens detection based on deep image
  representations,'' in \emph{Conference on Graphics, Patterns and Images
  (SIBGRAPI)}.\hskip 1em plus 0.5em minus 0.4em\relax IEEE, August 2015, pp.
  157--164.

\bibitem{Menotti:TIFS:2015}
D.~Menotti, G.~Chiachia, A.~Pinto, W.~R. Schwartz, H.~Pedrini, A.~X.
  Falc{\~{a}}o, and A.~Rocha, ``Deep representations for iris, face, and
  fingerprint spoofing detection,'' \emph{IEEE Transactions on Information
  Forensics and Security}, vol.~10, no.~4, pp. 864--879, April 2015.

\bibitem{Gragnaniello_SITIS_2016}
D.~Gragnaniello, C.~Sansone, G.~Poggi, and L.~Verdoliva, ``Biometric spoofing
  detection by a domain-aware convolutional neural network,'' in \emph{Int.
  Conference on Signal-Image Technology Internet-Based Systems (SITIS)}, Nov
  2016, pp. 193--198.

\bibitem{He_BTAS_2016}
L.~He, H.~Li, F.~Liu, N.~Liu, Z.~Sun, and Z.~He, ``Multi-patch convolution
  neural network for iris liveness detection,'' in \emph{IEEE Intl. Conference
  on Biometrics Theory, Applications and Systems (BTAS)}, September 2016, pp.
  1--7.

\bibitem{Raghavendra_WACV_2017}
R.~Raghavendra, K.~B. Raja, and C.~Busch, ``Contlensnet: Robust iris contact
  lens detection using deep convolutional neural networks,'' in \emph{{IEEE}
  Winter Conference on Applications of Computer Vision (WACV)}, March 2017, pp.
  1160--1167.

\bibitem{Lee_BS_2006}
S.~J. Lee, K.~R. Park, and J.~Kim, ``Robust fake iris detection based on
  variation of the reflectance ratio between the iris and the sclera,'' in
  \emph{Biometrics Symposium: Special Session on Research at the Biometric
  Consortium Conference}, September 2006, pp. 1--6.

\bibitem{Park_OptEng_2007}
\BIBentryALTinterwordspacing
J.~H. Park and M.~G. Kang, ``Multispectral iris authentication system against
  counterfeit attack using gradient-based image fusion,'' \emph{Optical
  Engineering}, vol.~46, no.~11, pp. 117\,003--117\,003--14, 2007. [Online].
  Available: \url{http://dx.doi.org/10.1117/1.2802367}
\BIBentrySTDinterwordspacing

\bibitem{Thavalengal_TCE_2016}
S.~Thavalengal, T.~Nedelcu, P.~Bigioi, and P.~Corcoran, ``Iris liveness
  detection for next generation smartphones,'' \emph{{IEEE} Trans. Consumer
  Electronics}, vol.~62, no.~2, pp. 95--102, May 2016.

\bibitem{Pacut_ICCST_2006}
A.~Pacut and A.~Czajka, ``Aliveness detection for iris biometrics,'' in
  \emph{{IEEE} Int. Carnahan Conference on Security Technology (ICCST)},
  October 2006, pp. 122--129.

\bibitem{Lee_IMA_2010}
\BIBentryALTinterwordspacing
E.~C. Lee and K.~R. Park, ``Fake iris detection based on {3D} structure of iris
  pattern,'' \emph{Intl. Journal of Imaging Systems and Technology}, vol.~20,
  no.~2, pp. 162--166, 2010. [Online]. Available:
  \url{http://dx.doi.org/10.1002/ima.20227}
\BIBentrySTDinterwordspacing

\bibitem{Connell_ASSP_2013}
J.~Connell, N.~Ratha, J.~Gentile, and R.~Bolle, ``Fake iris detection using
  structured light,'' in \emph{{IEEE} Int. Conference on Acoustics, Speech and
  Signal Processing}, May 2013, pp. 8692--8696.

\bibitem{Hughes_HICSS_2013}
K.~Hughes and K.~W. Bowyer, ``Detection of contact-lens-based iris biometric
  spoofs using stereo imaging,'' in \emph{Hawaii Intl. Conference on System
  Sciences}, January 2013, pp. 1763--1772.

\bibitem{Raja_TIFS_2015}
K.~Raja, R.~Raghavendra, and C.~Busch, ``Video presentation attack detection in
  visible spectrum iris recognition using magnified phase information,''
  \emph{{IEEE} Trans. Inf. Forens. Security}, vol.~10, no.~10, pp. 2048--2056,
  October 2015.

\bibitem{Rigas_PRL_2015}
\BIBentryALTinterwordspacing
I.~Rigas and O.~V. Komogortsev, ``Eye movement-driven defense against iris
  print-attacks,'' \emph{Pattern Recognition Letters}, vol. 68, Part 2, pp. 316
  -- 326, 2015, special Issue on “Soft Biometrics”. [Online]. Available:
  \url{http://www.sciencedirect.com/science/article/pii/S0167865515001737}
\BIBentrySTDinterwordspacing

\bibitem{Czajka_TIFS_2015}
A.~Czajka, ``Pupil dynamics for iris liveness detection,'' \emph{{IEEE} Trans.
  Inf. Forens. Security}, vol.~10, no.~4, pp. 726--735, April 2015.

\bibitem{Czajka_CSUR_2018}
\BIBentryALTinterwordspacing
A.~Czajka and K.~W. Bowyer, ``Presentation attack detection for iris
  recognition: An assessment of the state-of-the-art,'' \emph{ACM Comput.
  Surv.}, vol.~51, no.~4, pp. 86:1--86:35, Jul. 2018. [Online]. Available:
  \url{http://doi.acm.org/10.1145/3232849}
\BIBentrySTDinterwordspacing

\bibitem{Nguyen_2018}
\BIBentryALTinterwordspacing
D.~Nguyen, T.~Pham, Y.~Lee, and K.~Park, ``Deep learning-based enhanced
  presentation attack detection for iris recognition by combining features from
  local and global regions based on nir camera sensor,'' \emph{Sensors},
  vol.~18, no.~8, p. 2601, Aug 2018. [Online]. Available:
  \url{http://dx.doi.org/10.3390/s18082601}
\BIBentrySTDinterwordspacing

\bibitem{kontschieder2015deep}
P.~Kontschieder, M.~Fiterau, A.~Criminisi, and S.~Rota~Bulo, ``Deep neural
  decision forests,'' in \emph{{IEEE} Int. Conference on Computer Vision
  (ICCV)}, 2015, pp. 1467--1475.

\bibitem{ISO_30107-3_2017}
{ISO/IEC FDIS 30107-3:2017}, ``{Information technology -- Biometric
  presentation attack detection -- Part 3: Testing and reporting}.''

\bibitem{Yambay2014}
D.~Yambay, J.~S. Doyle, K.~W. Bowyer, A.~Czajka, and S.~Schuckers,
  ``Livdet-iris 2013 - iris liveness detection competition 2013,'' \emph{{IEEE}
  Int. Joint Conference on Biometrics (IJCB)}, pp. 1--8, 2014.

\bibitem{Yambay2015}
D.~Yambay, B.~Walczak, S.~Schuckers, and A.~Czajka, ``Livdet-iris 2015 - iris
  liveness detection competition 2015,'' \emph{IEEE Intl. Conference on
  Identity, Security and Behavior Analysis (ISBA)}, pp. 1--6, 2017.

\bibitem{kannala2012bsif}
J.~Kannala and E.~Rahtu, ``{BSIF}: Binarized statistical image features,'' in
  \emph{Pattern Recognition (ICPR), 2012 21st Intl. Conference on}.\hskip 1em
  plus 0.5em minus 0.4em\relax IEEE, 2012, pp. 1363--1366.

\bibitem{livdet2017}
\BIBentryALTinterwordspacing
``{LivDet-Iris 2017 -- Iris Liveness Detection Competition},'' Website, 2017,
  last access: 02/21/2018. [Online]. Available:
  \url{http://iris2017.livdet.org/}
\BIBentrySTDinterwordspacing

\bibitem{othman2016osiris}
\BIBentryALTinterwordspacing
N.~Othman, B.~Dorizzi, and S.~Garcia-Salicetti, ``{OSIRIS}: An open source iris
  recognition software,'' \emph{Pattern Recognition Letters}, vol.~82, no.~P2,
  pp. 124--131, Oct. 2016. [Online]. Available:
  \url{https://doi.org/10.1016/j.patrec.2015.09.002}
\BIBentrySTDinterwordspacing

\bibitem{ioffe2015batch}
S.~Ioffe and C.~Szegedy, ``Batch normalization: Accelerating deep network
  training by reducing internal covariate shift,'' in \emph{Intl. Conference on
  Machine Learning}, 2015, pp. 448--456.

\bibitem{kuncheva2014}
\BIBentryALTinterwordspacing
L.~I. Kuncheva, ``Combining label outputs,'' in \emph{Combining Pattern
  Classifiers}.\hskip 1em plus 0.5em minus 0.4em\relax John Wiley \& Sons,
  Inc., 2014, pp. 111--142. [Online]. Available:
  \url{http://dx.doi.org/10.1002/9781118914564.ch4}
\BIBentrySTDinterwordspacing

\bibitem{breiman2001random}
L.~Breiman, ``Random forests,'' \emph{Machine learning}, vol.~45, no.~1, pp.
  5--32, 2001.

\bibitem{genuer2010variable}
R.~Genuer, J.-M. Poggi, and C.~Tuleau-Malot, ``Variable selection using random
  forests,'' \emph{Pattern Recognition Letters}, vol.~31, no.~14, pp.
  2225--2236, 2010.

\bibitem{kittler1998combining}
J.~Kittler, M.~Hatef, R.~P. Duin, and J.~Matas, ``On combining classifiers,''
  \emph{IEEE Transactions on Pattern Analysis and Machine Intelligence},
  vol.~20, no.~3, pp. 226--239, 1998.

\bibitem{Moreno-Seco2006}
F.~Moreno-Seco, J.~M. I{\~{n}}esta, P.~J.~P. de~Le{\'o}n, and L.~Mic{\'o},
  ``Comparison of classifier fusion methods for classification in pattern
  recognition tasks,'' in \emph{Structural, Syntactic, and Statistical Pattern
  Recognition}, D.-Y. Yeung, J.~T. Kwok, A.~Fred, F.~Roli, and D.~de~Ridder,
  Eds.\hskip 1em plus 0.5em minus 0.4em\relax Berlin, Heidelberg: Springer
  Berlin Heidelberg, 2006, pp. 705--713.

\bibitem{Breiman:1984}
L.~Breiman, \emph{Classification and regression trees}, ser. Wadsworth
  statistics/probability series.\hskip 1em plus 0.5em minus 0.4em\relax
  Wadsworth International Group, 1984.

\bibitem{Fleiss:EPM:1973}
J.~L. Fleiss and J.~Cohen, ``The equivalence of weighted kappa and the
  intraclass correlation coefficient as measures of reliability,''
  \emph{Educational and Psychological Measurement}, vol.~33, no.~3, pp.
  613--619, 1973.

\bibitem{Doyle2015}
J.~S. Doyle and K.~W. Bowyer, ``Robust detection of textured contact lenses in
  iris recognition using {BSIF},'' \emph{IEEE Access}, vol.~3, pp. 1672--1683,
  2015.

\bibitem{Kohli2013}
N.~Kohli, D.~Yadav, M.~Vatsa, and R.~Singh, ``Revisiting iris recognition with
  color cosmetic contact lenses,'' \emph{Intl. Conference on Biometrics (ICB)},
  pp. 1--7, 2013.

\bibitem{Gupta2014}
P.~Gupta, S.~Behera, M.~Vatsa, and R.~Singh, ``On iris spoofing using print
  attack,'' \emph{Pattern Recognition (ICPR), 2014 22nd Intl. Conference on},
  pp. 1681--1686, 2014.

\bibitem{Yadav2014}
D.~Yadav, N.~Kohli, J.~S. Doyle, R.~Singh, M.~Vatsa, and K.~W. Bowyer,
  ``Unraveling the effect of textured contact lenses on iris recognition,''
  \emph{{IEEE} Trans. Inf. Forens. Security}, vol.~9, pp. 851--862, 2014.

\bibitem{Kohli2016}
N.~Kohli, D.~Yadav, M.~Vatsa, R.~Singh, and A.~Noore, ``Detecting medley of
  iris spoofing attacks using desist,'' \emph{{IEEE} Int. Conference on
  Biometrics: Theory Applications and Systems (BTAS)}, pp. 1--6, 2016.

\end{thebibliography}

\end{document}